\documentclass[5p]{elsarticle}

\usepackage{amssymb}
\usepackage{graphicx}
\usepackage{multirow}
\usepackage{caption}
\usepackage{color}
\usepackage{hyperref}
\usepackage{xcolor}
\usepackage{enumitem}
\usepackage{tabularx}
\usepackage{vcell}

\definecolor{mycitecolor}{RGB}{255,0,0} 
\hypersetup{
    colorlinks=true,
    linkcolor=mycitecolor,   
    citecolor=mycitecolor,   
    urlcolor=mycitecolor     
}

\begin{document}
\begin{frontmatter}



\title{Recent Advances in Deterministic Human Motion Prediction: A Review}


\author{Tenghao Deng}
\ead{dtenghao@shu.edu.cn}
\author{Yan Sun\corref{cor1}}
\ead{yansun@shu.edu.cn}


\address{School of Computer Engineering and Science, Shanghai University, Shanghai, China}

\cortext[cor1]{Corresponding author}
\begin{abstract}
In recent years, with the continuous advancement of deep learning and the emergence of large-scale human motion datasets, human motion prediction technology has gradually gained prominence in various fields such as human-computer interaction, autonomous driving, sports analysis, and personnel tracking. This article introduces common model architectures in this domain along with their respective advantages and disadvantages. It also systematically summarizes recent research innovations, focusing on in-depth discussions of relevant papers in these areas, thereby highlighting forward-looking insights into the field's development. Furthermore, this paper provides a comprehensive overview of existing methods, commonly used datasets, and evaluation metrics in this field. Finally, it discusses some of the current limitations in the field and proposes potential future research directions to address these challenges and promote further advancements in human motion prediction.
\end{abstract}




\end{frontmatter}


\section{Introducation}
\label{sec:intro}
\begin{figure*}
  \centering
  \includegraphics[width=1\textwidth]{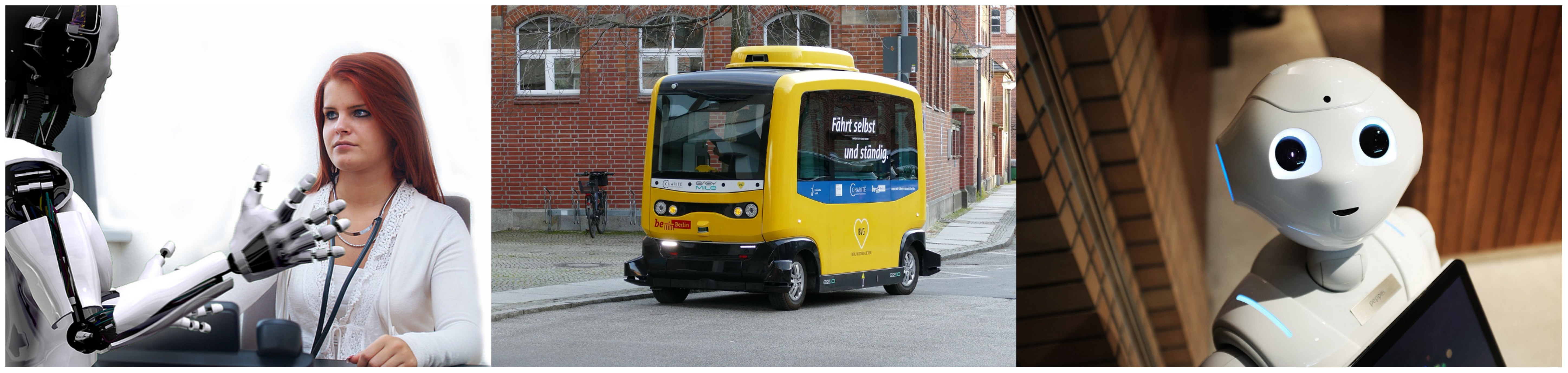}
  \caption{\textbf{The application domains of human body motion prediction}. On the left, it pertains to the field of human-machine interaction, where machines can respond appropriately based on human actions; in the middle, it extends to autonomous driving vehicles, which can quickly take evasive actions based on pedestrian motion trajectories; on the right, it encompasses intelligent robots that can comprehend human behaviors and provide enhanced services to humans.}
  \label{fig:1}
\end{figure*}

\begin{sloppypar}
Human motion prediction has gradually attracted widespread attention from researchers in recent years, with the rise of applications such as robotics, autonomous driving, and human-computer interaction\cite{padenSurveyMotionPlanning2016, koppulaAnticipatingHumanActivities2013, koppulaAnticipatingHumanActivities, lefkopoulosInteractionAwareMotionPrediction2021}, as shown in \hyperref[fig:1]{Fig.~\ref*{fig:1}}.
 Human motion prediction is to predict future possible actions based on the actions that have already occurred in the human body, which can deepen the machine’s understanding of the surrounding environment. For intelligent robots, human motion prediction can help them understand human behavior and coordinate interactive postures. For autonomous vehicles, it can recognize pedestrians on the road and predict whether they are preparing to cross the road or walk into the road, thereby better planning the driving path and avoiding potential dangers. However, human motion prediction may be an innate ability for humans, but it is a complex and challenging task for machines. How to accurately capture this spatio-temporal relationship from posture changes, and be able to cope with occlusions and incomplete data situations in the real world, and thus achieve accurate and robust prediction, there are still many challenges.
 \end{sloppypar}

\begin{sloppypar}
Early human motion predictions used traditional methods such as nonlinear Markov models\cite{brandStyleMachines2000, lehrmannEfficientNonlinearMarkov2014}, Gaussian process dynamic models\cite{wangGaussianProcessDynamical2005,urtasun3DPeopleTracking2006,wangGaussianProcessDynamical2008}, and restricted Boltzmann machines\cite{taylorModelingHumanMotion2006}. These methods mainly rely on statistics and probability theory. They often assume that data follows a certain specific distribution, but in fact the complexity and diversity of human actions make this assumption untenable. Therefore, they have shown effects in predicting simple actions, but they have encountered great problems in complex actions or long-term prediction problems. 

Thanks to the rapid development of deep learning in recent years, people have begun to use recurrent neural networks (RNNs) to extrapolate actions. RNNs that are often used to process sequence data seem very suitable for processing time series human action data. However, RNN has some inherent defects, such as gradient disappearance and gradient explosion problems\cite{usmanSkeletonbasedMotionPrediction2022}, which make it ineffective in processing long sequence data. Therefore, methods such as CNN and GAN are introduced. CNN can effectively extract spatial features through local receptive fields and weight sharing; while GAN can make the predicted future motion sequence more accurate and reasonable through adversarial training of generators and discriminators. In short, the field of human motion prediction has begun to flourish.
\end{sloppypar}

Although there are many different directions of research in the field of human motion prediction, this article will still limit the discussion scope to deterministic human motion prediction based on 3D skeletons, because skeleton data is usually more accurate and stable\cite{songStrongerFasterMore2020}, and deterministic human motion prediction is currently The strategy adopted by most research methods\cite{3DHumanMotion2022}, this limitation helps to make the discussion in this article more indepth and targeted. Moreover, this article takes a different approach from previous reviews by categorizing papers based on model types\cite{3DHumanMotion2022, marchellusDeepLearning3D2022, usmanSkeletonbasedMotionPrediction2022}, but instead, it summarizes six main innovative directions in recent years and discusses existing algorithms for human motion prediction based on these innovations. This approach allows us to provide a comprehensive overview of the technical details of existing methods and aids researchers gain a better understanding of the most recent developments and cutting-edge trends in the field of human motion prediction. By focusing on innovative directions, readers can more easily understand current research development trends and possible research directions and challenges. This approach makes this article more valuable and can provide a deep understanding and insight into the field of human motion prediction.

This article will be divided into several parts to explore related content in the field of human motion prediction. In the second part, we will clearly define the problem and discuss technical challenges. The third part will introduce various types of models involved, including RNNs, CNNs, GCNs, GANs etc. Subsequently, in the fourth part, we will discuss major innovative directions in recent years in depth including modifications to model structures; acquisition of global motion information; changes in input forms; auxiliary tasks etc., providing a comprehensive perspective on methods. In the fifth part we will further discuss datasets; evaluation metrics; comparison of method performance etc… Finally this article will look forward to future development trends in the field of human motion prediction providing some thoughts for readers’ future research.

Therefore, the main contributions of this paper can be summarized as follows:
\begin{enumerate}[label={\arabic*)}, itemsep=0pt, topsep=5pt, partopsep=0pt, parsep=0pt]
    \item In contrast to previous comprehensive reviews on the broader topic of human motion prediction, this paper is the first to provide a review specifically focused on deterministic human motion prediction.
    \item This paper provides a detailed exposition of the latest research methods, offering a comprehensive overview of their primary innovative directions from six distinct perspectives. This serves to facilitate researchers in gaining insight into the most recent trends within this field.
    \item This paper not only introduces the most common benchmark datasets and evaluation metrics in this field but also distinguishes itself from past reviews by employing a novel approach to comprehensively summarize the performance of numerous methods across various datasets. Lastly, we delve into an in-depth discussion of the ongoing challenges.
\end{enumerate}

\begin{figure}[h]
  \centering
  \includegraphics[width=0.4\textwidth]{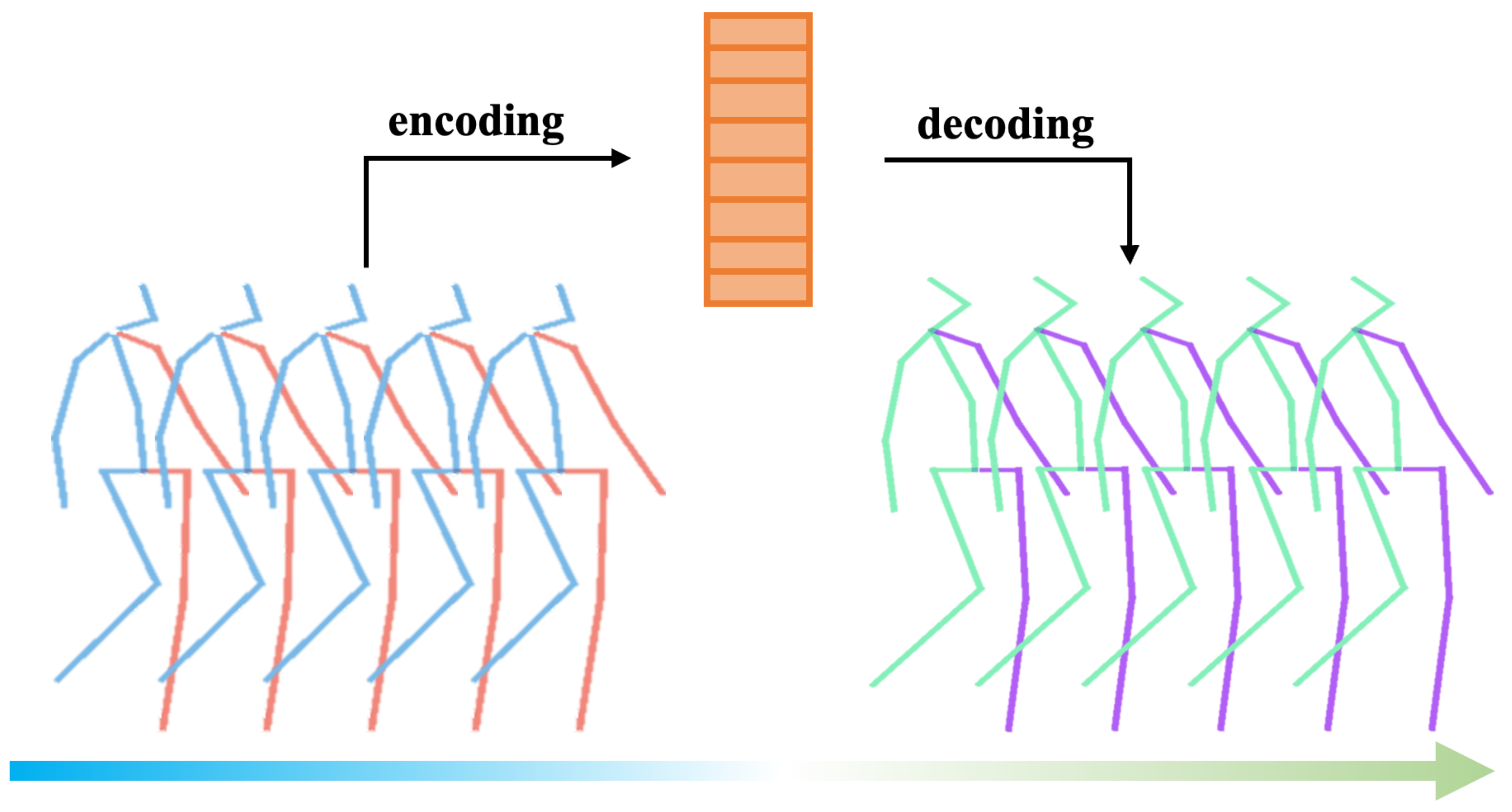}
  \caption{\textbf{Diagram for predicting human body motion}. The model's task is to capture and encode features from a historical sequence of motion and subsequently decode these features to make predictions.}
  \label{fig:2}
\end{figure}

\begin{figure*}[ht]
  \centering
  \includegraphics[width=0.8\textwidth]{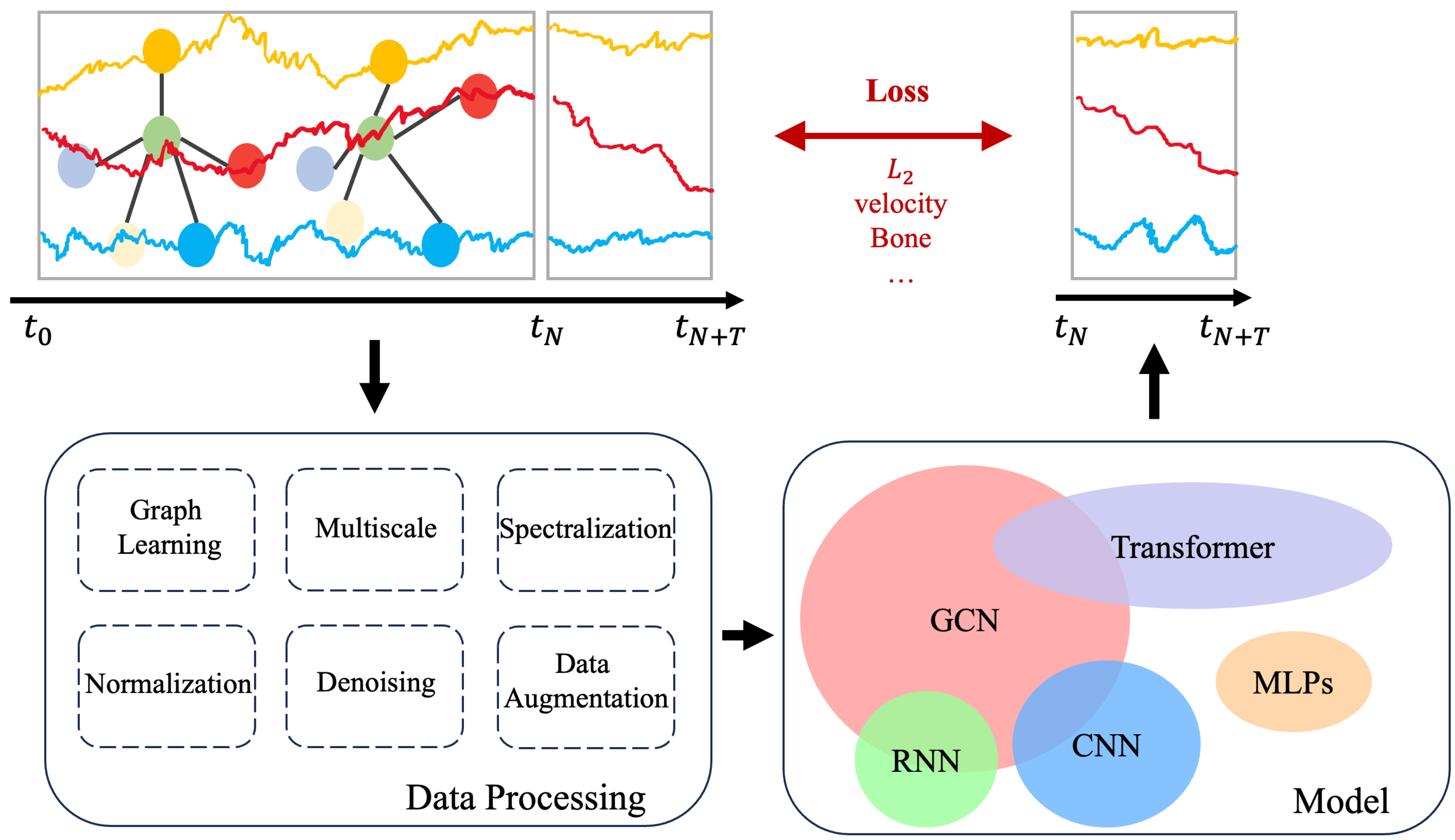}
  \caption{\textbf{Three stages of human motion prediction}: firstly, the data input and preprocessing stage; secondly, the model construction stage, in which various models such as recurrent neural networks (RNNs) and convolutional neural networks (CNNs) can be employed; and finally, the prediction output stage.}
  \label{fig:3}
\end{figure*}

\section{Problem formulation and challenges}
\label{sec:Problem}
Before the formal discussion of methods, a clear definition of the deterministic human motion prediction problem based on skeletons needs to be made. Human posture is expressed by representative joints\cite{Human36MLarge, vonmarcardRecoveringAccurate3D2018, mahmoodAMASSArchiveMotion2019} , so 3D skeleton human motion is a sequence of human postures arranged in time order. From a mathematical point of view, an action sequence of length N can be defined as $X_{1:N}=(x_1,\ldots,x_N)\in\mathbb{R}^{N\times J\times D}$, where $J$ represents the number of joints needed to form a human posture, and $D$ represents the number of features of each joint. Therefore, $x_i \in \mathbb{R}^{J \times D}$, where $i \in (1, N)$, represents the posture of the $i$th frame. The future action sequence to be predicted can be represented as $X_{N+1:N+T}=(x_{N+1},\ldots,x_{N+T})\in\mathbb{R}^{T\times J\times D}$, where $T$ represents the number of frames to be predicted. \hyperref[fig:2]{Fig.~\ref*{fig:2}} illustrates the basic human motion prediction problem. The core task of human motion prediction is to predict the future action sequence $X_{N+1:N+T}$ based on the existing historical action sequence $X_{1:N}$, which can be represented by the prediction model $F$. Therefore, the prediction formula can be written as $\widehat{X}_{N+1:N+T}=F(X_{1:N})$, where $\widehat{X}_{N+1:N+T}$ represents the future action sequence predicted by the model, and our goal is to make $\widehat{X}_{N+1:N+T}$ as close as possible to the true value $X_{N+1:N+T}$.

The human motion prediction problem poses many challenges, such as difficulty in accurately capturing complex temporal relationships in human motion, changes in actions, long-term dependencies, and modeling complex spatial posture relationships between joints. In long-term prediction, even the currently most advanced methods\cite{gaoDecomposeMoreAggregate} will tend to average posture due to the diversity of actions, which deviates seriously from actual situations. In practical applications, it is also necessary to cope with challenges such as occlusion and incomplete data, which further increases the complexity of the problem. Moreover, human motion prediction needs to be real-time, which means that the prediction model must be able to process input data in real time or near real time and generate accurate motion predictions. This is a capability that most current models do not have. However, with the continuous efforts of researchers, more and more innovative solutions have been proposed. This article will discuss them in the following sections.

\section{Network structure design}
\label{sec:Network}
As shown in \hyperref[fig:3]{Fig.~\ref*{fig:3}}, human motion prediction can be divided into three stages, which are data processing, model, and output. In order to better capture the complex spatio-temporal relationships between joints, the design of the network structure is extremely critical. From the earliest Recurrent Neural Networks (RNNs) to the introduction of Convolutional Neural Networks (CNNs) and Generative Adversarial Networks (GANs), each model has its unique advantages and disadvantages. Therefore, extensive and diversified research has been conducted in the field of human motion prediction to extract effective features to enhance the performance of motion prediction. In this section, we will briefly introduce the methods according to different model types, so that readers can understand the application of various models in solving human motion prediction problems. The selection and design of these models are of great significance for improving prediction accuracy, enhancing robustness, and achieving real-time performance.

\subsection{RNN}
\begin{sloppypar}
Due to the fact that action prediction is often regarded as a sequence-to-sequence forecasting task, recurrent neural networks (RNNs) have been widely adopted due to their commendable performance in such tasks. 
\begin{figure}[ht]
  \centering
  \includegraphics[width=0.4\textwidth]{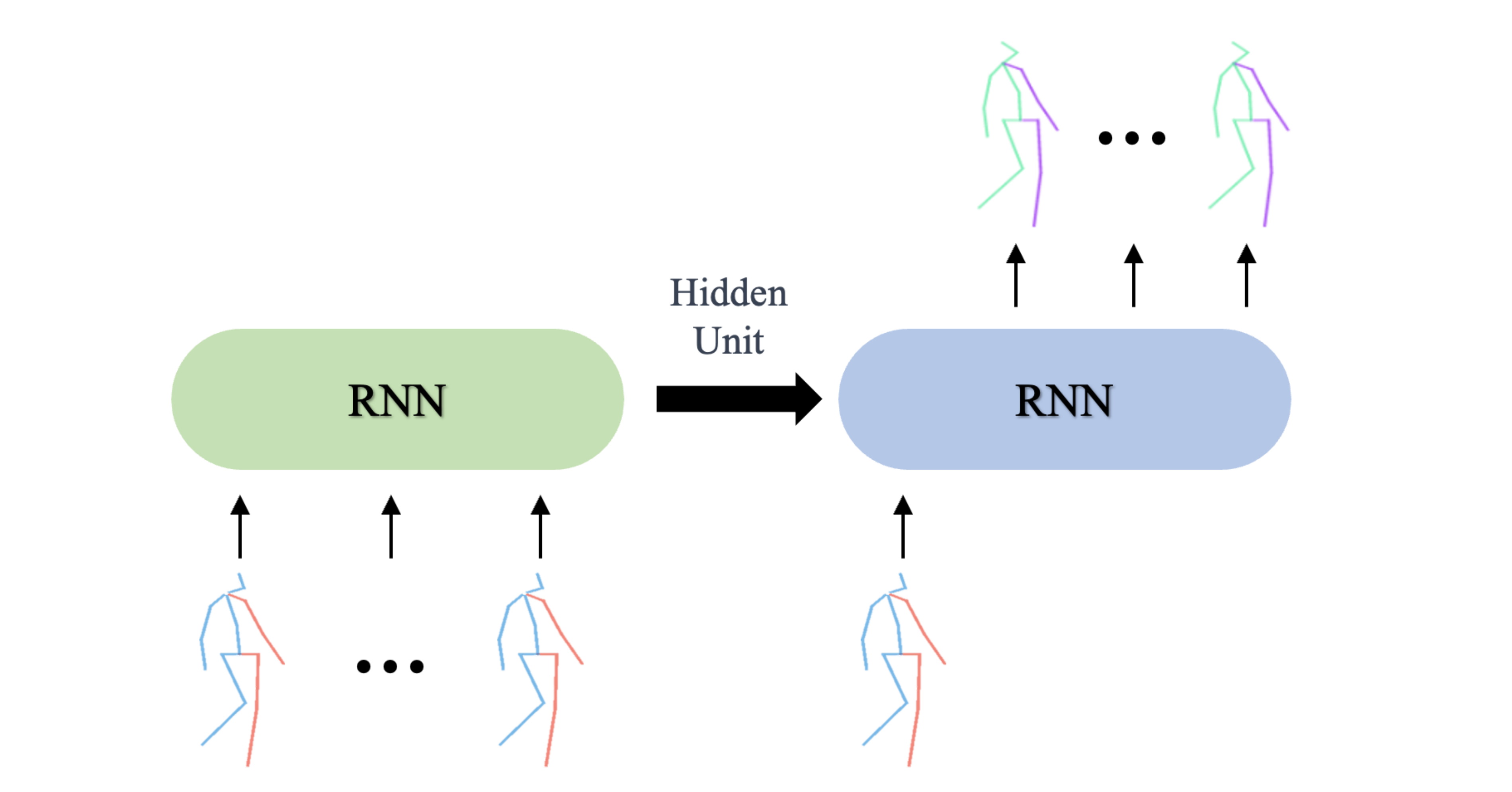}
  \caption{\textbf{The fundamental framework for RNN prediction}  consists of the following steps: firstly, encoding the historical action sequence, and subsequently transmitting the encoded information in the form of hidden variables to the decoder. The decoder, relying on the previous time step's posture and hidden variables, generates the posture prediction for the current time step through a recursive process.}
  \label{fig:4}
\end{figure} 
Fragkiadaki et al.\cite{fragkiadakiRecurrentNetworkModels2015} introduced an encoder-decoder framework that embeds human poses into a latent space and subsequently employs Long Short-Term Memory networks (LSTMs) for continuous updates and predictions. However, it is evident that \cite{fragkiadakiRecurrentNetworkModels2015} did not adequately consider the spatial dependencies among joints. Consequently, Jain et al.\cite{jainStructuralRNNDeepLearning2016} explicitly designed spatial correlations between hands, feet, and the torso, thus better modeling the spatial dependencies of poses. Tang et al.\cite{tangLongTermHumanMotion2018} obtained spatial information by passing pose features through fully connected layers and aggregated the historical pose information as the dependent variable, while Liu et al.\cite{liuNaturalAccurateFuture2019} explicitly modeled joint-level spatial relationships using LSTMs. In addition to modeling spatial dependencies, Martinez et al.\cite{martinezHumanMotionPrediction2017} also introduced residual connections to enable RNNs to better model angular velocities rather than just absolute values. 

\begin{figure}[b]
  \centering
  \includegraphics[width=0.5 \textwidth]{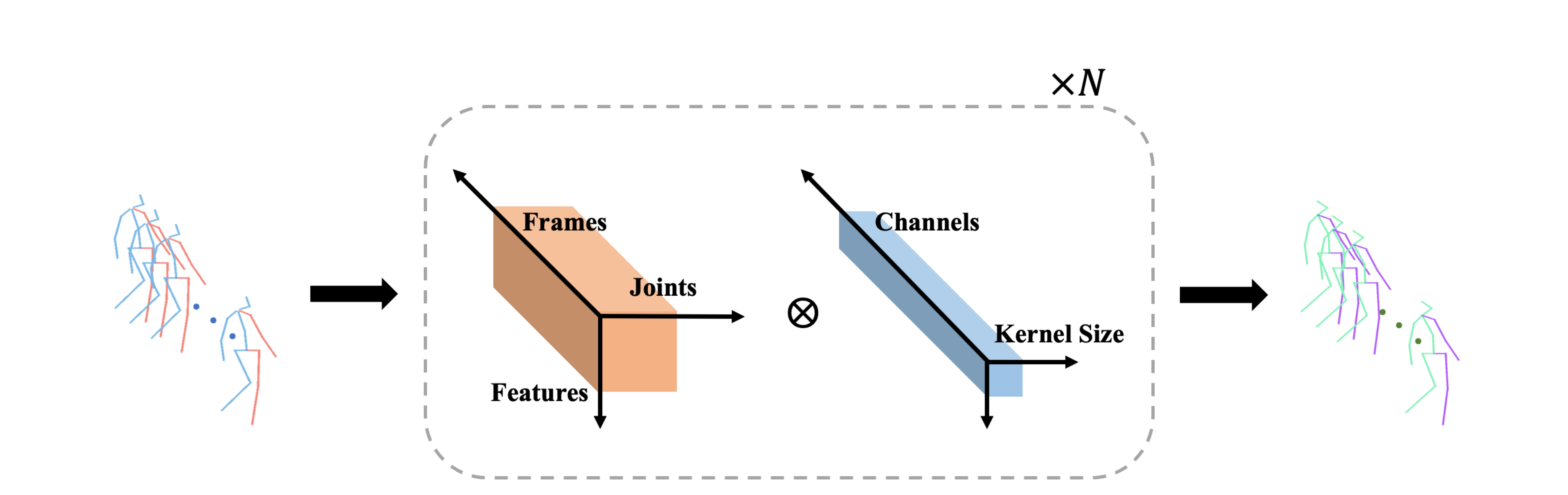}
  \caption{\textbf{The fundamental framework of CNN prediction}  involves representing the human body's posture in a form akin to an image, followed by the application of Convolutional Neural Networks (CNN) for processing.}
  \label{fig:5}
\end{figure}

Despite the aforementioned methods addressing the spatiotemporal relationships of actions to some extent, there still exist several inherent limitations in RNN models. For instance, difficulty in processing spatial information\cite{shuSpatiotemporalCoattentionRecurrent2019, caetanoSkeleMotionNewRepresentation2019a}, sequential computation makes it challenging to parallelize training and inference\cite{pengRWKVReinventingRNNs2023}, handling longer historical sequences often leads to gradient vanishing or exploding issues\cite{marchellusDeepLearning3D2022}, prediction errors accumulate over time\cite{uedaSpatiotemporalAggregationSkeletal2022}, resulting in poses eventually converging to the mean\cite{tangLongTermHumanMotion2018} . Although some methods \cite{butepageDeepRepresentationLearning2017, butepageAnticipatingManyFutures2018, chiuActionAgnosticHumanPose2019, liuNaturalAccurateFuture2019, liIndependentlyRecurrentNeural2018} 
attempt to mitigate these issues, the predictive structure inherent to RNNs, as illustrated in \hyperref[fig:4]{Fig.~\ref*{fig:4}}, continues to restrict their effectiveness in alleviating training difficulty, error accumulation, discontinuous predictions, and poor long-term forecasting performance.
\end{sloppypar}

\subsection{CNN}
\begin{sloppypar}
Convolutional Neural Networks (CNNs) are typically used to process image data, and due to their excellent ability to capture spatial dependencies, more and more researchers are beginning to adopt CNN architectures to handle human motion prediction problems. As shown in \hyperref[fig:5]{Fig.~\ref*{fig:5}}, the fundamental model structure involves transforming historical human motion sequences into a format resembling images, followed by multiple applications of 2D convolution operations for analysis. Li et al.\cite{liConvolutionalSequenceSequence2018} adopted a multi-layer convolutional architecture, where the lower layers are used to capture dependencies between nearby frames, and the higher layers are used to capture dependencies between distant frames, thus encoding the sequence into long-term and short-term features. This bears some resemblance to the use of CNNs in the field of image processing, where shallow convolutional layers capture finer details, such as edges and textures, while deeper convolutional layers capture more abstract and global features, such as object shapes and structures. Liu et al.\cite{liuTrajectoryNetNewSpatiotemporal2020} fully utilized the advantages of CNNs, treating sequence trajectories as channels to jointly process spatio-temporal information and globally model temporal co-occurrence information. Omar et al.\cite{medjaouriHRSTANHighResolutionSpatioTemporal2022} adopted a high-resolution network architecture to obtain different granular observation patterns and introduced a novel high-resolution spatio-temporal attention mechanism that enhances spatio-temporal information in feature maps without increasing computational complexity.
\end{sloppypar}

In addition to conventional CNN architectures, there is a form called Temporal Convolutional Network (TCN), which is often used to replace Recurrent Neural Networks (RNNs) to learn temporal information of sequences. For example, Theodoros et al. \cite{sofianosSpaceTimeSeparableGraphConvolutional2021} used TCN to further integrate temporal information and make predictions, while Kareem et al.\cite{eltounyDETGNUncertaintyAwareHuman2023} introduced a large number of TCN structures in their ensemble model to better aggregate time dependencies. Although CNNs have excellent time feature aggregation capabilities, the fixed convolution kernel size can also limit the model's performance in some cases. Therefore, researchers are constantly exploring how to further optimize and improve CNN architectures to better adapt to the needs of human motion prediction tasks. These innovations will provide more possibilities for improving the performance and accuracy of motion prediction.

\subsection{GCN}
\begin{sloppypar}
Strictly speaking, Graph Convolutional Networks (GCNs) are a form of Convolutional Neural Networks (CNNs), but their object is graph data. Nevertheless, we will set up a separate section for GCNs because they have received widespread attention and research in the field of human motion prediction. Mao et al.\cite{maoLearningTrajectoryDependencies2020} modeled human posture as a graph structure and used a learnable GCN to automatically connect joints without manual intervention. Their approach enables the system to automatically learn the connections between joints, thereby enhancing the accuracy and robustness of posture modeling. Cui et al.\cite{cuiLearningDynamicRelationships2020} believed that unconstrained learning could lead to unstable training, so they divided the adjacency matrix into natural connections and hidden connections, and found that the introduction of natural connections could improve model performance. While previous methods either used GCNs to simultaneously process spatio-temporal information or only processed spatial information. Theodoros et al.\cite{sofianosSpaceTimeSeparableGraphConvolutional2021} proposed the first spatio-temporally separable GCN, applying GCNs independently to spatial information and temporal information processing, significantly reducing the number of parameters and improving performance. Subsequently, a large number of research works on GCNs emerged, including multi-scale graph networks\cite{liDynamicMultiscaleGraph2020}, equivariant graph networks\cite{xuEqMotionEquivariantMultiagent2023}, graph scattering networks\cite{liSkeletonPartedGraphScattering2022} , multi-task graph networks\cite{cuiAccurate3DHuman2021} , spatio-temporal gated graph networks\cite{zhongSpatioTemporalGatingAdjacencyGCN2022}, etc.
\end{sloppypar}

However, some researchers also found that as the network depth increases, graph convolution may cause features between joints to become similar\cite{chenSimpleDeepGraph2020}. Therefore, some research, such as \cite{liDirectedAcyclicGraph2021}, tried to limit the direction of feature aggregation by using directed graphs and mutual aggregation between joints and skeletons, while \cite{caoDualAttentionModel2022} extended GCN into a multi-layer residual graph structure to increase the diversity of joint features. Although GCNs also have some shortcomings, they are currently still one of the most mainstream models in the field of human motion prediction. The wide application and continuous improvement of GCNs have brought many interesting research directions and opportunities to this field.

\subsection{Transformer}
As the attention mechanism introduced by the Transformer has excelled in various fields such as natural language and image processing, in recent years, numerous researchers have begun incorporating this model into the domain of human motion prediction. Aksan et al.\cite{aksanSpatiotemporalTransformer3D2021} employed attention mechanisms to calculate the correlation between joints and temporal frames. However, the use of frame-based pose attention may lead to ambiguous motion predictions, as static poses fail to provide information about the direction of motion. Subsequently, Mao et al.\cite{maoMultilevelMotionAttention2021} divided historical motion sequences into multiple sub-sequences and discovered periodic motion information by computing the similarity between the current sub-sequence and historical sub-sequences. Cai et al.\cite{LearningProgressiveJoint} utilized the Transformer architecture to encode features from historical motion sequences, during decoding, they employed a progressive prediction strategy to predict joints gradually from the center to the periphery based on skeletal structure. This approach fully harnesses the motion chains of the skeletal structure, which are highly significant in human motion, and integral to human behaviors. Beyond attention mechanism applications, researchers have also explored modifications to the Transformer structure, making the model more suitable for human motion prediction tasks. For example, some work \cite{zhongGeometricAlgebrabasedMultiview2023, caoDualAttentionModel2022} enhanced the representation of Query and Key in the attention mechanism. In addition, some recent research \cite{medjaouriHRSTANHighResolutionSpatioTemporal2022, cuiMetaAuxiliaryLearningAdaptive2023, xuHumanMotionPrediction2023} innovated the entire Transformer framework to adapt to the specific requirements of human motion prediction tasks. In summary, as research on Transformer models continues to advance, an increasing number of improvements and variants are proposed for applications in human motion prediction, providing us with more powerful and accurate methods to address this task.

\subsection{MLP}
Multilayer Perceptron (MLP), as one of the longstanding models in the history of deep learning, has traditionally been considered more suitable for processing unstructured data and seemed less adept at handling highly structured human pose data. However, in a surprising turn of events in 2022, researchers such as Arij et al.\cite{bouaziziMotionMixerMLPbased3D2022} made a significant discovery. They found that, compared to models relying on skeletal structure priors and computationally expensive convolutional networks like GCN, models based on MLPs exhibit lower computational complexity and are better at capturing spatial and temporal dependencies in body posture. Subsequently, Guo et al.\cite{guoBackMLPSimple} also adopted MLP models. In contrast to \cite{bouaziziMotionMixerMLPbased3D2022}, they did not use squeeze-and-excitation blocks and activation layers, relying solely on fully connected layers, transpose layers, and layer normalization. Nevertheless, they achieved unexpectedly impressive results, This method can deliver results comparable to more complex approaches such as GCN or Transformer, while offering faster execution and a smaller model size. Wang et al.\cite{wangMixerLayerWorth2023} further attempted to introduce MLPs into their GCN model to leverage the strengths of both methods for improved extraction of spatiotemporal dependencies. This suggests that human motion prediction tasks can be modeled in a novel and simplified manner without the explicit fusion of spatial and temporal information. These studies provide crucial inspiration for future research endeavors.

\subsection{GAN}
\begin{table*}
\centering
\caption{Comparative analysis of motion prediction models}
\label{table:1}
\begin{tabular}{lp{6.2cm}p{6.2cm}ll} 
\hline
Model Type  & Advantages                                                                                            & Disadvantages                                                                              & Related Work                                                                                                                                            \\ 
\hline
RNN         & Handles time-series data, suitable for continuous motion trajectories.                                & Issues with vanishing or exploding gradients, error accumulation in long-term predictions. & \cite{liuInvestigatingPoseRepresentations2023,dongSkeletonBasedHumanMotion2022}                                                                     \\
CNN         & Excellent temporal feature aggregation.                                                               & Limited spatial scale generalization due to fixed kernel sizes.                            & \cite{medjaouriHRSTANHighResolutionSpatioTemporal2022,tangTemporalConsistencyTwostream2022,liuTrajectoryNetNewSpatiotemporal2020}                   \\
GCN         & Processes graph-structured data, considers skeletal connections and human relationships.              & Feature similarity between joints with increasing network depth.                           & \cite{gaoDecomposeMoreAggregate,zhongGeometricAlgebrabasedMultiview2023,wangSpatioTemporalBranchingMotion2023,xuEqMotionEquivariantMultiagent2023}  \\
Transformer & Powerful sequence modeling, parallel processing of sequences, suitable for variable-length sequences. & Requires substantial data for training, complex model.                                     & \cite{cuiMetaAuxiliaryLearningAdaptive2023,daiKDFormerKinematicDynamic2023,xuHumanMotionPrediction2023}                                             \\
MLP         & Simple and easy to implement, effective for non-structural features.                                  & Poor performance for complex motion patterns, struggles with long-term dependencies.       & \cite{guoBackMLPSimple,bouaziziMotionMixerMLPbased3D2022}                                                                                           \\
GAN         & Generates realistic human motion sequences                                                            & Unstable training, requires meticulous hyperparameter tuning.                              & \cite{zhaoBidirectionalTransformerGAN2023,cuiLearningDynamicRelationships2020}                                                                      \\
\hline
\end{tabular}
\end{table*}

Inspired by the mechanism of adversarial training, some research works have begun to introduce discriminators to assist the learning of neural networks. For example, Cui et al.\cite{cuiLearningDynamicRelationships2020} introduced a discriminator and adversarial loss, aiming to make the sequence generated by the generator closer to the real posture sequence. On the other hand, Zhao et al.\cite{zhaoBidirectionalTransformerGAN2023} adopted a dual discriminator, including a sequence-based discriminator and a frame-based discriminator. The tasks of these two discriminators are to distinguish the predicted sequence from the real sequence from both local frame and global sequence perspectives. Although theoretically, Generative Adversarial Network (GAN) models have many potential advantages, in practice it is difficult to achieve a balance between the generator and the discriminator, so the training process is somewhat difficult.

In conclusion, various models in the field of human motion prediction exhibit distinct characteristics. A profound understanding of these attributes is crucial for the selection and design of future methods. Therefore, we will summarize the strengths and weaknesses of different models in the domain of human motion prediction in \hyperref[table:1]{Table~\ref{table:1}} for reference. 

\begin{figure}
  \centering
  \includegraphics[width=0.4\textwidth]{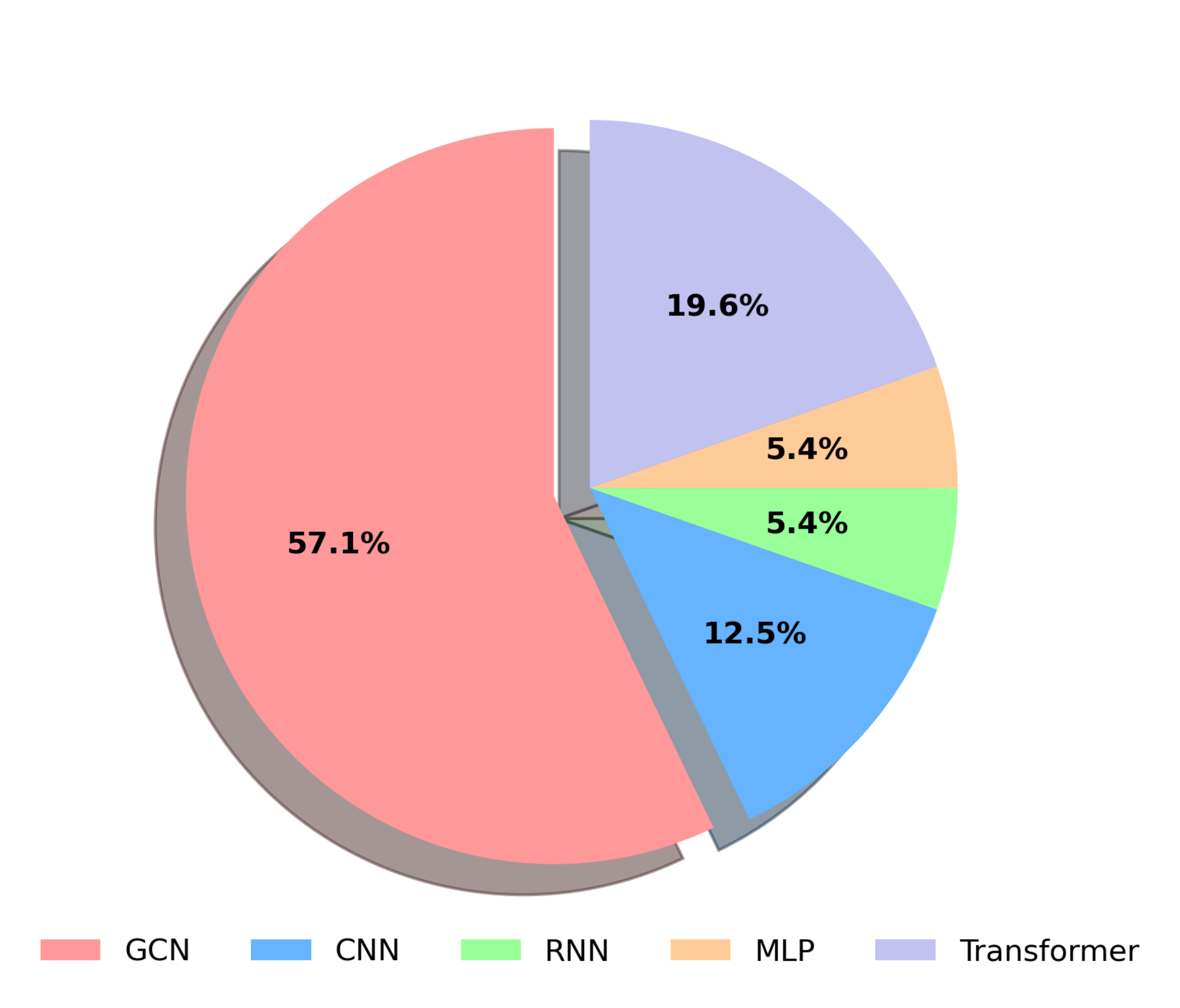}
  \caption{The distribution of model types used in relevant research since 2020.}
  \label{fig:10}
\end{figure}

Additionally, we conducted a statistical analysis of the model types utilized in papers published since 2020, along with their respective publication years, as summarized in \hyperref[table:2]{Table~\ref{table:2}}. It is evident that Graph Convolutional Networks (GCNs) have emerged as one of the most widely applied model types in recent years, featuring prominently in numerous research papers, particularly since 2020. GCN continues to be a pivotal model of interest among researchers. Concurrently, Transformer models have also received significant attention, particularly in 2022 and 2023, with several research papers adopting this model. Transformers have already demonstrated remarkable performance in domains such as natural language processing and computer vision, and their popularity is growing in research areas such as time series data processing and others. Although certain studies still employ other model types, such as Convolutional Neural Networks (CNNs), Recurrent Neural Networks (RNNs), and Multilayer Perceptrons (MLPs), their usage is comparatively less frequent. After visual analysis, as shown in  \hyperref[fig:10]{Fig.~\ref*{fig:10}}, the results are highly intuitive. Among them, more than half of the papers employed GCN models, indicating that using GCNs for capturing spatiotemporal dependencies may be the current mainstream research direction.

\begin{table}
\centering
\small
\caption{Model type and publication year of the paper.}
\begin{tabular}{ccc} 
\hline
Methods        & Categories                   & Years  \\ 
\hline
DMAB\cite{gaoDecomposeMoreAggregate}           & \multirow{28}{*}{GCN}        & 2023   \\
GA-MIN\cite{zhongGeometricAlgebrabasedMultiview2023}         &                              & 2023  \\
STB-GCN\cite{wangSpatioTemporalBranchingMotion2023}        &                              & 2023   \\
EqMotion\cite{xuEqMotionEquivariantMultiagent2023}       &                              & 2023   \\
IT-GCN\cite{sunAccurateHumanMotion2023}         &                              & 2023   \\
GcnMlpMixer\cite{wangMixerLayerWorth2023}    &                              & 2023   \\
ResChunk\cite{zandMultiscaleResidualLearning2023}    &                              & 2023   \\
SANet\cite{heInitialPredictionFinetuning2023}          &                              & 2023   \\
PatternGCN\cite{tangCollaborativeMultidynamicPattern2023}     &                              & 2023   \\
CGHMP\cite{liClassguidedHumanMotion2023}          &                              & 2023   \\
LSM-GCN\cite{wangLearningSnippettoMotionProgression2023}        &                              & 2023   \\
DS-GCN\cite{fuLearningConstrainedDynamic2023}         &                              & 2023   \\
SC-GCN\cite{chenSpatiotemporalConsistencyLearning2023}         &                              & 2023   \\
FMS-AM\cite{fernandoRememberingWhatImportant2023}         &                              & 2023   \\
Multi-GCN\cite{renMultiGraphConvolutionNetwork2023}      &                              & 2023   \\
DANet\cite{caoDualAttentionModel2022}          &                              & 2022   \\
SPGSN\cite{liSkeletonPartedGraphScattering2022}          &                              & 2022   \\
PGBIG\cite{maProgressivelyGeneratingBetter2022}          &                              & 2022   \\
MT-GCN\cite{cuiAccurate3DHuman2021}         &                              & 2022   \\
PK-GCN\cite{sunOverlookedPosesActually2022}         &                              & 2022   \\
LTM-GCN\cite{kicirogluLongTermMotion2022a}        &                              & 2022   \\
MultiAttention\cite{maoMultilevelMotionAttention2021} &                              & 2021   \\
STS-GCN\cite{sofianosSpaceTimeSeparableGraphConvolutional2021}        &                              & 2021   \\
MSR-GCN\cite{dangMSRGCNMultiScaleResidual}        &                              & 2021   \\
TC-GCN\cite{liuMotionPredictionUsing}         &                              & 2021   \\
DirGCN\cite{liDirectedAcyclicGraph2021}         &                              & 2021   \\
LDR-GCN\cite{cuiLearningDynamicRelationships2020}        &                              & 2020   \\
TrajGCN\cite{maoLearningTrajectoryDependencies2020}        &                              & 2020   \\
\hline
MAL-Attention\cite{cuiMetaAuxiliaryLearningAdaptive2023}  & \multirow{10}{*}{Transformer} & 2023   \\
AuxFormer\cite{xuAuxiliaryTasksBenefit2023}     &                              & 2023   \\

KD-Former\cite{daiKDFormerKinematicDynamic2023}      &                              & 2023   \\
SmtpFormer\cite{xuHumanMotionPrediction2023}     &                              & 2023   \\
BiAttention\cite{zhaoBidirectionalTransformerGAN2023}    &                              & 2023   \\
StaAttention\cite{uedaSpatiotemporalAggregationSkeletal2022}   &                              & 2022   \\
CH-TR\cite{mascaroRobustHumanMotion2022}          &                              & 2022   \\
POTR\cite{martinez-gonzalezPoseTransformersPOTR}           &                              & 2021   \\
STAttention\cite{aksanSpatiotemporalTransformer3D2021}    &                              & 2021   \\
LPAttention\cite{LearningProgressiveJoint}    &                              & 2020   \\ 
\hline
HR-STAN\cite{medjaouriHRSTANHighResolutionSpatioTemporal2022}        & \multirow{4}{*}{CNN}         & 2022   \\
TC-CNN\cite{tangTemporalConsistencyTwostream2022}         &                              & 2022   \\
TrajectoryNet\cite{liuTrajectoryNetNewSpatiotemporal2020}  &                              & 2020   \\
TIM-GCN\cite{lebaillyMotionPredictionUsing2021}        &                              & 2020   \\ 
\hline
AHMR\cite{liuInvestigatingPoseRepresentations2023}           & \multirow{2}{*}{RNN}         & 2022   \\
PVRED\cite{wangPVREDPositionVelocityRecurrent2021}             &                              & 2021   \\ 
PS\cite{dongSkeletonBasedHumanMotion2022}             &                              & 2020   \\ 
\hline
siMLPe\cite{guoBackMLPSimple}         & \multirow{2}{*}{MLP}         & 2023   \\
MotionMixer\cite{bouaziziMotionMixerMLPbased3D2022}    &                              & 2022   \\ 
\hline
SPOTR\cite{nargundSPOTRSpatiotemporalPose2023}          & GCN,transformer              & 2023   \\
MSTP-Net\cite{chenMSTPNetMultiscaleSpatiotemporal2023}       & GCN,CNN,transformer          & 2023   \\
DE-TGN\cite{eltounyDETGNUncertaintyAwareHuman2023}         & GCN,CNN                      & 2023   \\
CoGNN\cite{liOnlineMultiAgentForecasting2021}          & GCN,MLP                      & 2022   \\
mNAT\cite{liMultitaskNonAutoregressiveModel2021}           & GCN,CNN                      & 2020   \\
DMGNN\cite{liDynamicMultiscaleGraph2020}          & GCN,RNN                      & 2020   \\
\hline
\label{table:2}
\end{tabular}
\end{table}

\section{Recent advances in deterministic human motion prediction}
\begin{figure*}
  \centering
  \includegraphics[width=1\textwidth]{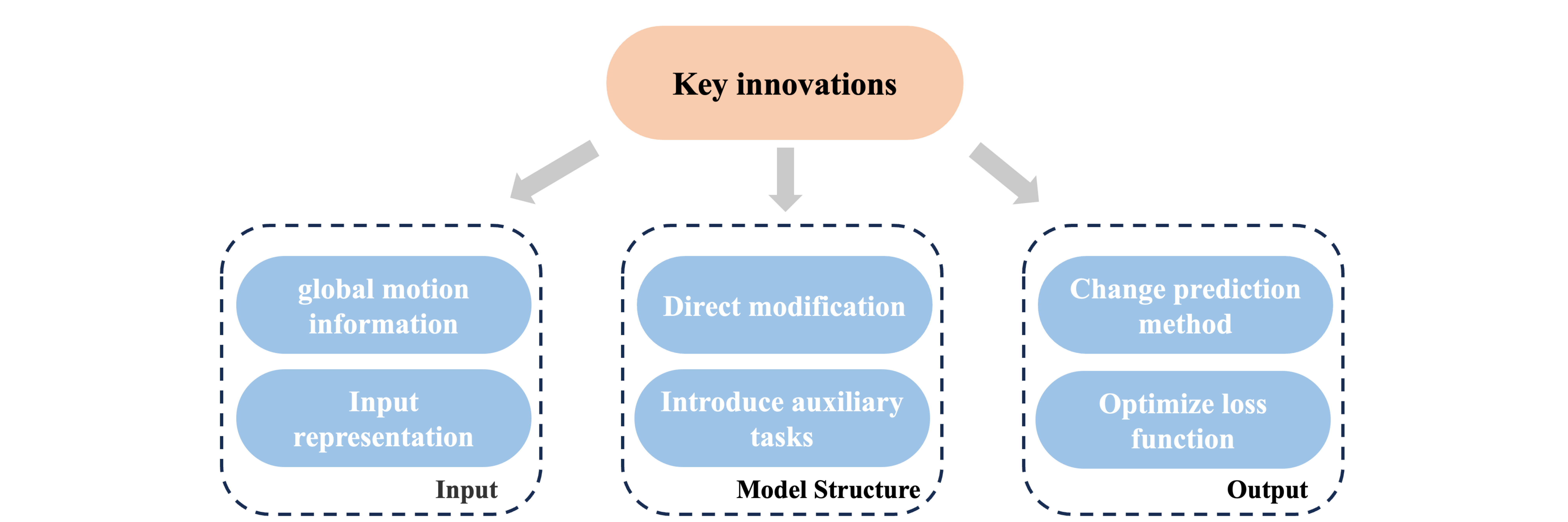}
  \caption{The primary direction of innovation.}
  \label{fig:6}
\end{figure*}

In this section, we will delve into the main innovative directions in recent years within the field of human motion prediction. As illustrated in \hyperref[fig:6]{Fig.~\ref*{fig:6}}, we categorize and discuss these directions based on three stages: input, model, and output. These directions primarily encompass improvements in model architecture, capture of global motion information, variations in input data, introduction of auxiliary tasks, alterations in prediction methods, and optimization of loss functions, among others. These innovations not only expand the application scope but also provide new ways for enhancing prediction accuracy and stability. By thoroughly exploring these innovative directions, this paper will showcase the latest advancements in the field of human motion prediction.

\subsection{Model structure}
One of the primary directions of innovation lies in the structural advancements of models. While there are numerous aspects to explore, due to space constraints and the focus of this paper on research conducted after 2020, we will concentrate specifically on structural innovations related to GCN and Transformer models.

Graph convolutional models are currently the mainstream models for 3D skeleton human motion prediction, and a large number of works are improving the structure of graph convolution, hoping to better model spatio-temporal dependencies. Zhong et al.\cite{zhongGeometricAlgebrabasedMultiview2023} used the concept of multivectors in geometric algebra to decompose the features $H$ and the transformation matrix $W$ in graph convolution. This approach enables the reconstruction of features from a spectral perspective, resulting in the construction of more robust graph feature representations. Cao et al.\cite{caoDualAttentionModel2022} introduced multi-layer residual graph convolution to overcome the smoothness problem brought by graph convolution. Xu et al.\cite{xuEqMotionEquivariantMultiagent2023} believed that simply transforming the input sequence into an abstract feature vector would lose the geometric relationship between motions, so they designed special equivariant operations to ensure the equivariance of motion, allowing the model to learn feature representations associated with geometric transformations. This enhances the model's ability to generalize across motion data with different geometric transformations. Wang et al.\cite{wangMixerLayerWorth2023} first used MLP to process structure-unknown dependencies, and then used GCN to process structure-known data. This adeptly combines the strengths of both models, resulting in an enhancement of the method's performance. Tang et al.\cite{tangCollaborativeMultidynamicPattern2023} used symmetric residual connections to replace traditional equidistant residual connections, making the distance between input and output closer. This, to some extent, mitigates the feature smoothing issue caused by GCN.

\begin{sloppypar}
Ma et al.\cite{maProgressivelyGeneratingBetter2022} used dense graph convolution as a basic unit and found that dense graph convolution performed better than spatio-temporal graph convolution. Some work\cite{wangLearningSnippettoMotionProgression2023, cuiLearningDynamicRelationships2020, liuMotionPredictionUsing} changed the representation of the adjacency matrix and represented the adjacency matrix in forms such as skeleton connections, motion prior knowledge, and learnable hidden connections. 
\begin{figure}
  \centering
  \includegraphics[width=0.5\textwidth]{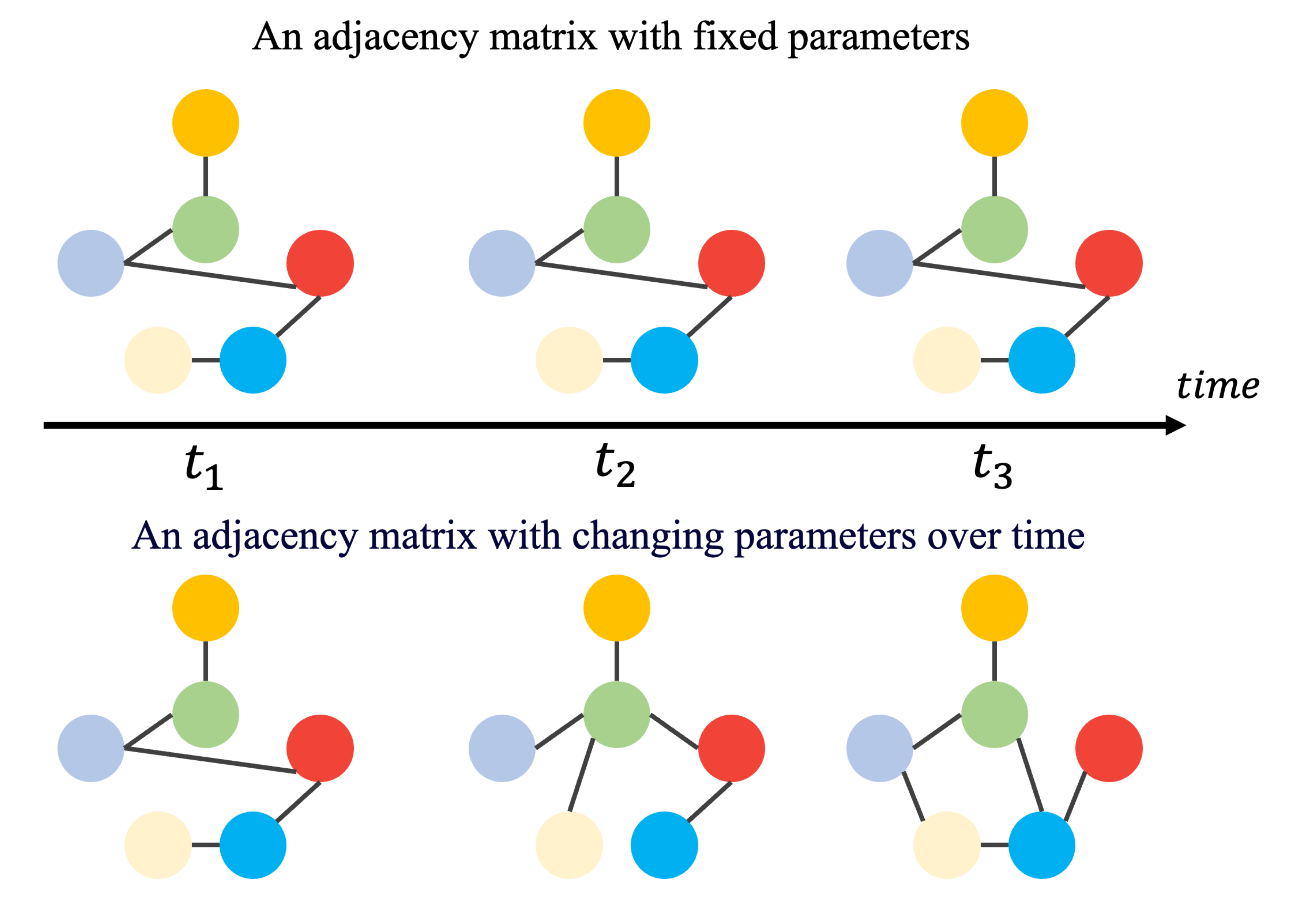}
  \caption{The difference between 'spatiotemporal-unshared' and 'spatiotemporal-shared'.}
  \label{fig:12}
\end{figure}
This approach better captures the relationships between skeletons and the prior information of motion, thereby providing richer contextual information. Ren et al.\cite{renMultiGraphConvolutionNetwork2023} represented human posture as a sparse graph and used multiple adjacency matrices to deal with joints at different distances separately, concluding that it is not the case that the larger the receptive field, the better the model performs, and also stated that natural connections play a crucial role in action prediction. Some work\cite{fuLearningConstrainedDynamic2023, zhongSpatioTemporalGatingAdjacencyGCN2022}  propose an adaptive adjacency matrix that can adjust according to the differences in samples, exhibiting a sample-specific characteristic. Notably, the adjacency matrix in \cite{fuLearningConstrainedDynamic2023} possesses a spatiotemporal-unshared feature, meaning it changes over time. This is similar to the spatiotemporal separable graph convolution proposed by Theodoros et al.\cite{sofianosSpaceTimeSeparableGraphConvolutional2021}. \hyperref[fig:12]{Fig.~\ref*{fig:12}} briefly illustrates the difference between spatiotemporal-unshared and spatiotemporal-shared. However, unlike \cite{sofianosSpaceTimeSeparableGraphConvolutional2021}, \cite{fuLearningConstrainedDynamic2023} uses a constrained adjacency matrix to represent general correlations and adjusts the correlations between each frame and joint based on the input-output dynamic correlations. This provides a new perspective for addressing the challenges brought about by differences between samples. These studies continuously improve graph convolution models, optimize structures, adjacency matrices, etc., from multiple angles to more accurately and powerfully model spatio-temporal dependencies.
\end{sloppypar}

Inspired by recent innovations in deep learning, attention mechanisms have received widespread attention. Researchers have tried to introduce attention mechanisms into models in order to more accurately focus on key joints or sequence segments, thereby enhancing perception of important parts of actions. The power of Query and Key representations is critical for attention model performance, zhong et al.\cite{zhongGeometricAlgebrabasedMultiview2023} used a structure similar to an hourglass network to enhance Query and Key in order to obtain a more compact spatial representation, while Cao et al.\cite{caoDualAttentionModel2022} used local-global dual attention to enhance Query and Key representation. There are also papers focusing on the lightweighting of attention models, such as omar et al.\cite{medjaouriHRSTANHighResolutionSpatioTemporal2022} adopted Efficient Attention proposed by \cite{shenEfficientAttentionAttention2021}, significantly reducing the spatio-temporal complexity of attention mechanisms. Cui et al.\cite{cuiMetaAuxiliaryLearningAdaptive2023} introduced sparse relay Transformer which can not only capture local neighbor joint features but also capture long-term relationships. 
Xu et al.\cite{xuHumanMotionPrediction2023}, modified SmtpFormer based on MetaFormer as basic unit, to handle spatio-temporal dependencies based on the conclusion: MetaFormer universal architecture is key for Transformer and MLP-like models to achieve excellent results in visual tasks \cite{yuMetaFormerActuallyWhat2022}.
 These studies not only introduce attention mechanisms into models and accurately focus on key joints and sequence segments but also further enhance model's spatio-temporal perception and prediction performance through different structural changes and combinations. \hyperref[table:3]{Table~\ref{table:3}} provides a comprehensive summary of these methods.

\begin{table*}[ht]
\centering
\caption{Methods for modifications to Model Structure.}
\label{table:3}
\begin{tabular}{lp{8.3cm}lll} 
\hline
Method                                                        & Insights                                                                                                                                                                       & Venue                                                                           & Year & Code                                                     \\ 
\hline
DMAB\cite{zhongGeometricAlgebrabasedMultiview2023}            & Learning features over multiple frequency spaces by varying the sampling window size and sampling interval                                                                       & CVPR                                                                            & 2023 & No                                                       \\
EqMotion\cite{xuEqMotionEquivariantMultiagent2023}            & To ensure the equivariance of motion, specialized equivariant operations are employed                                                                                            & CVPR                                                                            & 2023 & \href{https://github.com/MediaBrain-SJTU/EqMotion}{Yes}  \\
PatternGCN\cite{tangCollaborativeMultidynamicPattern2023}     & Employing TIM to handle temporal dependencies (assigning greater weight to more recent frames); Substituting symmetric residual connections for equidistant residual connections & IEEE~TCSVT                                                                      & 2023 & No                                                       \\
Multi-GCN\cite{renMultiGraphConvolutionNetwork2023}           & Utilizing adjacency matrices with joints spaced at intervals of 1, 2, 3, and so on                                                                                               & arXiv                                                                           & 2023 & No                                                       \\
DS-GCN\cite{fuLearningConstrainedDynamic2023}                 & Representing the adjacency matrix in a generic and dynamic form that can be adaptively adjusted based on the variations in input samples.                                        & IEEE TNNLS                                                                      & 2023 & \href{https://github.com/Jaakk0F/DSTD-GCN}{Yes}          \\
GA-MIN\cite{zhongGeometricAlgebrabasedMultiview2023}          & Passing Q and K through a hourglass network~to obtain a compact spatial representation                                                                                           & \begin{tabular}[c]{@{}l@{}}Pattern Recognition\textit{}\\\textit{}\end{tabular} & 2023 & No                                                       \\
MAL-Attention\cite{cuiMetaAuxiliaryLearningAdaptive2023}      & Utilizing a sparse relay transformer to reduce model complexity                                                                                                                  & arXiv                                                                           & 2023 & No                                                       \\
SmtpFormer\cite{xuHumanMotionPrediction2023}                  & Replacing the Attention module with a straightforward Temporal pooling                                                                                                           & ICIPMV                                                                          & 2023 & No                                                       \\
DANet\cite{caoDualAttentionModel2022}                         & Dual attention was used, with local attention to Q and K alone, followed by global attention to~Q, K, and V                                                                      & Neurocomputing                                                                  & 2022 & \href{https://github.com/lishuangshuang2022/DANet}{Yes}  \\
HR-STAN\cite{medjaouriHRSTANHighResolutionSpatioTemporal2022} & Efficient Attention is employed as a replacement for traditional Transformers                                                                                                    & CVPR                                                                            & 2022 & No                                                       \\
PGBIG\cite{maProgressivelyGeneratingBetter2022}               & Processing Spatio-Temporal Information with S-DGCN and T-DGCN                                                                                                                    & CVPR                                                                            & 2022 & \href{https://github.com/705062791/PGBIG}{Yes}           \\
LDR-GCN\cite{cuiLearningDynamicRelationships2020}             & Splitting the graph convolution into real connection adjacency matrices and a global graph                                                                                       & CVPR                                                                            & 2020 & No                                                       \\
\hline
\end{tabular}
\end{table*}

\begin{table*}
\centering
\caption{Methods for incorporating Global Motion Information.}
\label{table:4}
\begin{tabular}{lp{11cm}llll} 
\hline
Method                                                & Insights                                                                                                                       & Venue & Year & Code                                                      \\ 
\hline
IT-GCN\cite{sunAccurateHumanMotion2023}               & To obtain long-term aggregated features, an attention mechanism is employed                                                      & arXiv & 2023 & No                                                        \\
EqMotion\cite{xuEqMotionEquivariantMultiagent2023}    & To learn the pattern features of each joint, it remains invariant to euclidean transformations                                   & CVPR  & 2023 & \href{https://github.com/MediaBrain-SJTU/EqMotion}{Yes}   \\
FMS-AM\cite{fernandoRememberingWhatImportant2023}     & To use memory to store subject/task-specific knowledge                                                                           & arXiv & 2023 & No                                                        \\
MultiAttention\cite{maoMultilevelMotionAttention2021} & To employ an attention mechanism to capture the similarity between the current action context and historical action subsequences & IJCV  & 2021 & \href{https://github.com/wei-mao-2019/HisRepItself}{Yes}  \\
POTR\cite{martinez-gonzalezPoseTransformersPOTR}      & When using a transformer for encoding, an additional token is introduced to learn category information                           & ICCV  & 2021 & \href{https://github.com/idiap/potr}{Yes}                 \\
LPAttention\cite{LearningProgressiveJoint}            & To employ a dictionary to learn certain motion patterns of the joints                                                            & ECCV  & 2020 & No                                                        \\
\hline
\end{tabular}
\end{table*}

\subsection{Global motion information}
According to the research in literature \cite{sunAccurateHumanMotion2023}, directly dealing with excessively long motion history in the frequency domain may have a negative impact on model performance. This is because the model has difficulty capturing minor motion details, leading to a decrease in prediction accuracy. 
However, for complex motions that require a longer time to fully display their features, a longer input motion history is necessary 
\begin{figure}[hb]
  \centering
  \includegraphics[width=0.5\textwidth]{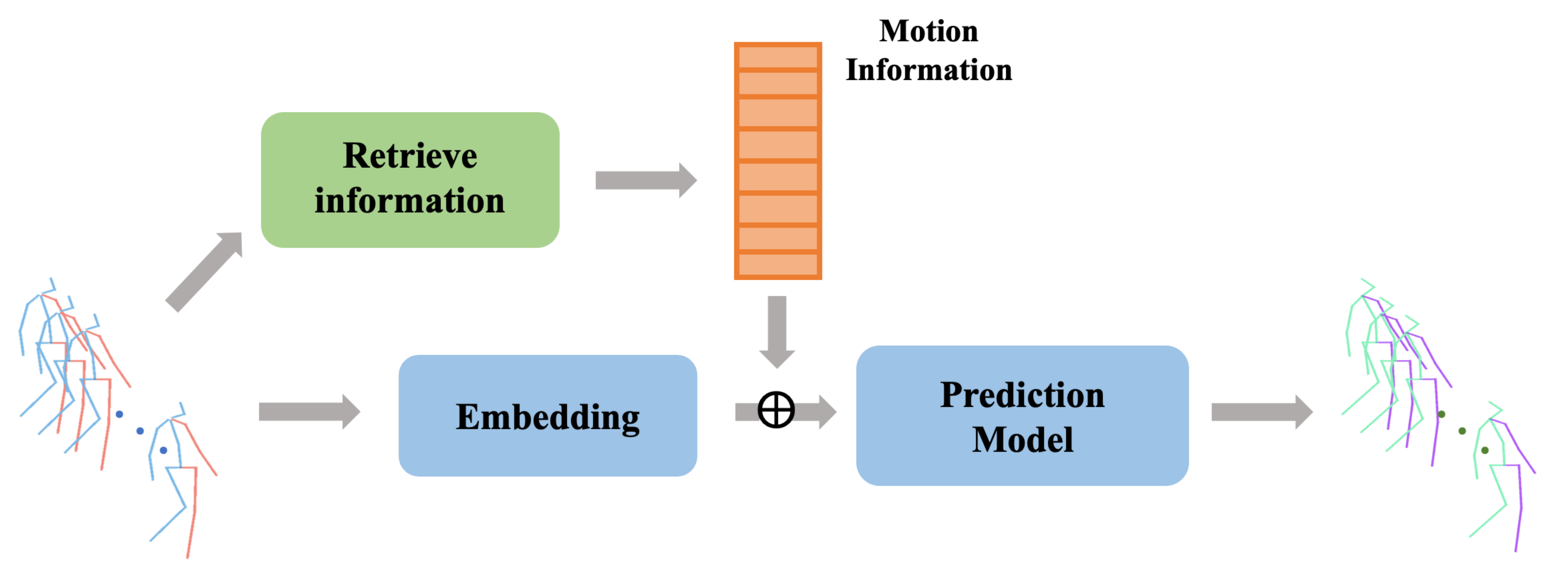}
  \caption{\textbf{Using global motion information to help predict}. The green block represents the global information extractor, which functions to incorporate global motion information into the original features, aiding the model in achieving a more comprehensive understanding of motion sequences.}
  \label{fig:7}
\end{figure}
because it is difficult to accurately grasp the future motion direction by relying solely on the movement of joint coordinates for prediction. As show in \hyperref[fig:7]{Fig.~\ref*{fig:7}}, if the model can obtain global information about the motion in advance, such as the type or pattern of motion, then under the guidance of global information, the accuracy of prediction will be greatly improved. Therefore, Xu et al.\cite{xuEqMotionEquivariantMultiagent2023} used an individual branch to calculate motion pattern features and used these features to guide the extraction of equivariant features. Mao et al.\cite{maoMultilevelMotionAttention2021} introduced attention mechanisms to calculate the similarity between the latest subsequence and historical subsequences, thereby summarizing information from historical sequences, which greatly improved the prediction accuracy of cyclical motions. Angel et al.\cite{martinez-gonzalezPoseTransformersPOTR} introduced additional tokens into the Transformer model and used them to capture global information of the sequence.
\begin{figure}[ht]
  \centering
  \includegraphics[width=0.4\textwidth]{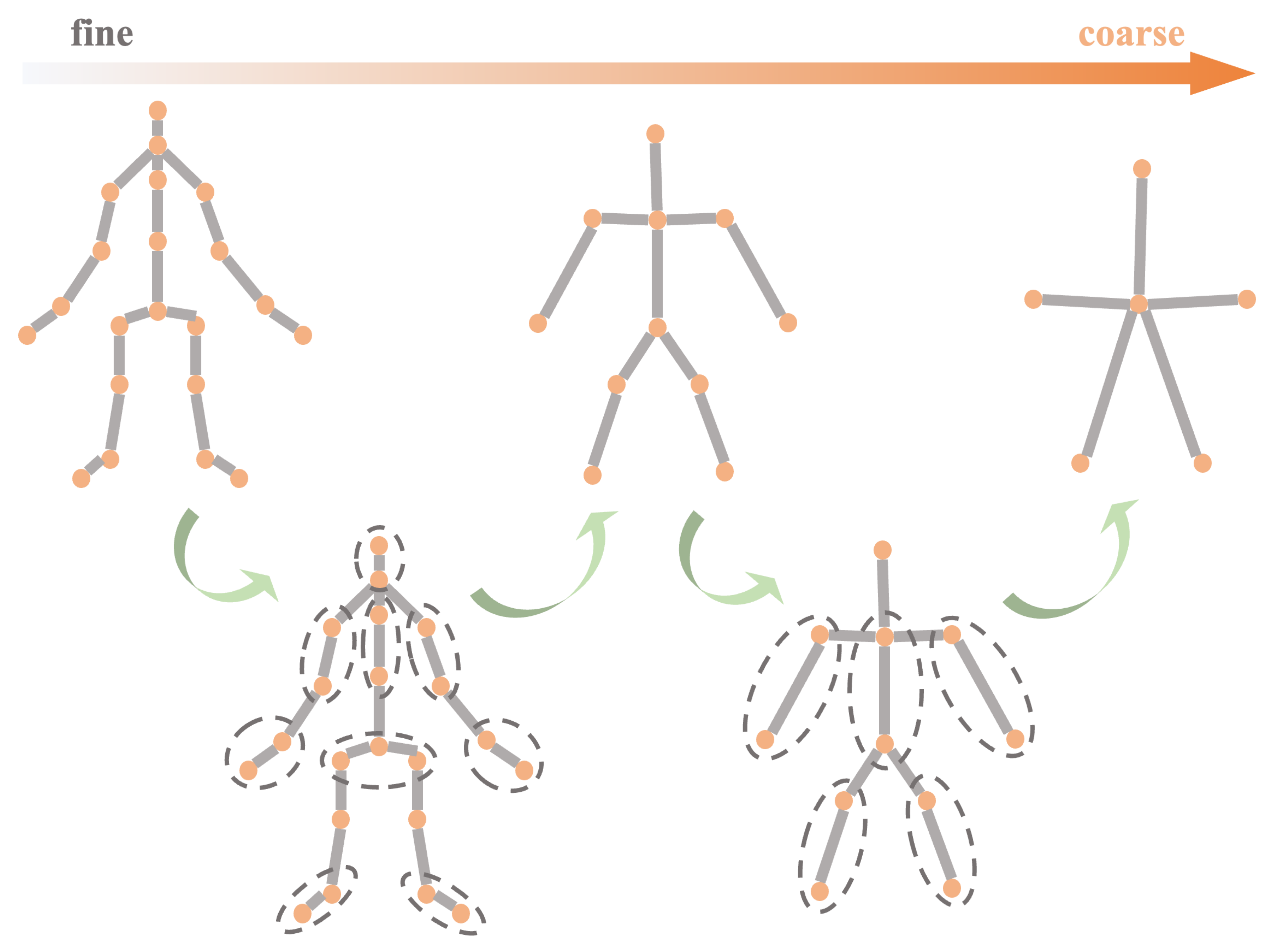}
  \caption{\textbf{Different fine-grained representations of pose}. From the left side to the right side of the diagram, adjacent joints continuously merge, with the granularity gradually becoming coarser}
  \label{fig:8}
\end{figure}

On the other hand, Cai et al.\cite{LearningProgressiveJoint} used a learnable dictionary to store pattern information for all motions. Tharindu et al.\cite{fernandoRememberingWhatImportant2023} used a dynamic masking method to extract character and action type features from each motion sequence. This approach ingeniously accounts for the differences in motion characteristics between various individuals and actions, thus contributing valuable insights to the research. A large number of research results also clearly show that for different types of complex motions, by calculating motion pattern features, long-time sequence motion histories can be effectively handled, thereby improving prediction accuracy. The review and summary of these methods can be found in \hyperref[table:4]{Table~\ref{table:4}}.

\subsection{Input representations}
When researchers deal with input action sequences, 
data is usually presented in the form of 3D coordinates or angles. However, to enhance the understanding ability of the network and effectively extract features, researchers can inject some prior knowledge into the input data. This approach helps to optimize the performance of action prediction. A common method is to use a dual-stream network, where one stream inputs velocity information and the other inputs position information. Tang et al.\cite{tangTemporalConsistencyTwostream2022} found that velocity information helps short-term prediction, while position information helps long-term prediction. Similarly, Chen et al.\cite{chenSpatiotemporalConsistencyLearning2023} adopted a dual-stream structure, but they sampled frame-by-frame on one stream and cross-frame on the other to integrate information at different time scales. Other methods such as Itsuki et al.\cite{uedaSpatiotemporalAggregationSkeletal2022} choose to use position as encoder input and angle as decoder input. This strategy facilitates the model in capturing spatial arrangement and directional information more effectively in the context of motion. Li et al.\cite{liDynamicMultiscaleGraph2020} use differential operations and use position, velocity, and acceleration as inputs at the same time. This approach naturally introduces dynamical characteristics, mitigates the discontinuity in pose generation, and predicts smoother sequences of human motion. While Li et al.\cite{liuInvestigatingPoseRepresentations2023} use quaternion as input because it has continuity and avoids discontinuity and singularity.

In addition, changing the granularity of the input or dividing the posture into multiple parts is also a common strategy. Multi-scale input plays an important role in human motion prediction, can capture details and overall information at the same time, adapt to motions of different scales, alleviate information loss, enhance robustness, and better deal with challenges such as occlusion and noise. For example, Mao et al.\cite{maoMultilevelMotionAttention2021} use whole body, partial body (trunk, left arm, right arm, left leg, right leg), joints as input, while Li et al.\cite{liSkeletonPartedGraphScattering2022} simply divide the body into upper body and lower body, both achieving good results. Tang et al.\cite{tangCollaborativeMultidynamicPattern2023} divided it into static, non-active, active three modes according to the activity level of joints and processed them separately. This approach allows for explicit utilization of the relationships between joints exhibiting the same dynamic mode, while avoiding unnecessary trajectory constraints under a global pattern.

\begin{table*}
\centering
\caption{Methods for modifying Input Representations.}
\label{table:5}
\begin{tabular}{lp{10cm}llll} 
\hline
Method & Insights & Venue & Year & Code \\ 
\hline
ResChunk\cite{zandMultiscaleResidualLearning2023} & Utilizing Multilayer Perceptrons (MLPs) to learn the correlations between joints and obtaining coarse-grained pose information through clustering techniques & arXiv & 2023 & \href{https://github.com/MohsenZand/ResChunk}{Yes} \\
PatternGCN\cite{tangCollaborativeMultidynamicPattern2023} & Dividing the joints into three states: static, inactive, and active, and processing them separately using GCN & IEEE TCSVT & 2023 & No \\
SC-GCN\cite{chenSpatiotemporalConsistencyLearning2023} & To alleviate the challenge of capturing dynamic contextual information by employing velocity representations & IEEE TCSVT & 2023 & No \\
AHMR\cite{liuInvestigatingPoseRepresentations2023} & Using the Stiefel manifold as input to bring about global continuity and reduce computational complexity when optimizing the loss & IEEE TPAMI & 2022 & \href{https://github.com/BII-wushuang/Lie-Group-Motion-Prediction}{Yes} \\
SPGSN\cite{liSkeletonPartedGraphScattering2022} & Dividing the body into multiple parts (upper and lower) & ECCV & 2022 & \href{https://github.com/MediaBrain-SJTU/SPGSN}{Yes} \\
TC-CNN\cite{tangTemporalConsistencyTwostream2022} & Dual-stream network, with one stream taking velocity as input, and the other taking position as input & Neurocomputing & 2022 & \href{https://github.com/BUPTJinZhang/TCTS_CNN}{Yes} \\
StaAttention\cite{uedaSpatiotemporalAggregationSkeletal2022} & Using position as the input for the encoder and angle as the input for the decoder & Array & 2022 & No \\
MSR-GCN\cite{dangMSRGCNMultiScaleResidual} & Perform graph convolution operations on pose scales from fine to coarse, and then from coarse to fine & \textcolor[rgb]{0.204,0.286,0.369}{CVPR} & 2021 & \href{https://github.com/Droliven/MSRGCN}{Yes} \\
DMGNN\cite{liDynamicMultiscaleGraph2020} & Learning features at different scales separately & \textcolor[rgb]{0.204,0.286,0.369}{CVPR} & 2020 & \href{https://github.com/limaosen0/DMGNN}{Yes} \\
\hline
\end{tabular}
\end{table*}

As shown in \hyperref[fig:8]{Fig.~\ref*{fig:8}}, in addition to categorizing the joints, more comprehensive and stable motion information can be captured based on different granularities, Coarse-grained representations, as compared to fine-grained ones, typically exhibit smaller changes during motion, greater stability in motion patterns, and are easier to predict. Dang et al.\cite{dangMSRGCNMultiScaleResidual} adopted a stage-by-stage prediction method that scales the body from fine to coarse and then from coarse to fine to gradually improve prediction results. Li et al.\cite{liDynamicMultiscaleGraph2020} used parallel input of three different granularities of posture to integrate information at different scales. The choice of these methods depends on specific tasks and research objectives and can be flexibly applied as needed to improve the accuracy and robustness of action prediction.

However, Zand et al.\cite{zandMultiscaleResidualLearning2023} have proposed a different perspective. They argue that traditional multiscale methods often rely on fixed structural prior information to aggregate proximal joint information, which may not adapt flexibly to different motion patterns. In real-world scenarios, different motion patterns can lead to strong correlations between joints, and even distant joints can influence each other. Therefore, Zand et al.\cite{zandMultiscaleResidualLearning2023} introduced a novel approach where they utilized multilayer perceptrons (MLPs) to learn inter-joint relationships and obtained coarse-grained pose information through clustering techniques. This approach exhibits a high level of insightfulness and adaptability. \hyperref[table:5]{Table~\ref{table:5}} offers a summary of the methods mentioned above.

\subsection{Auxiliary tasks}
\begin{figure}
  \centering
  \includegraphics[width=0.5\textwidth]{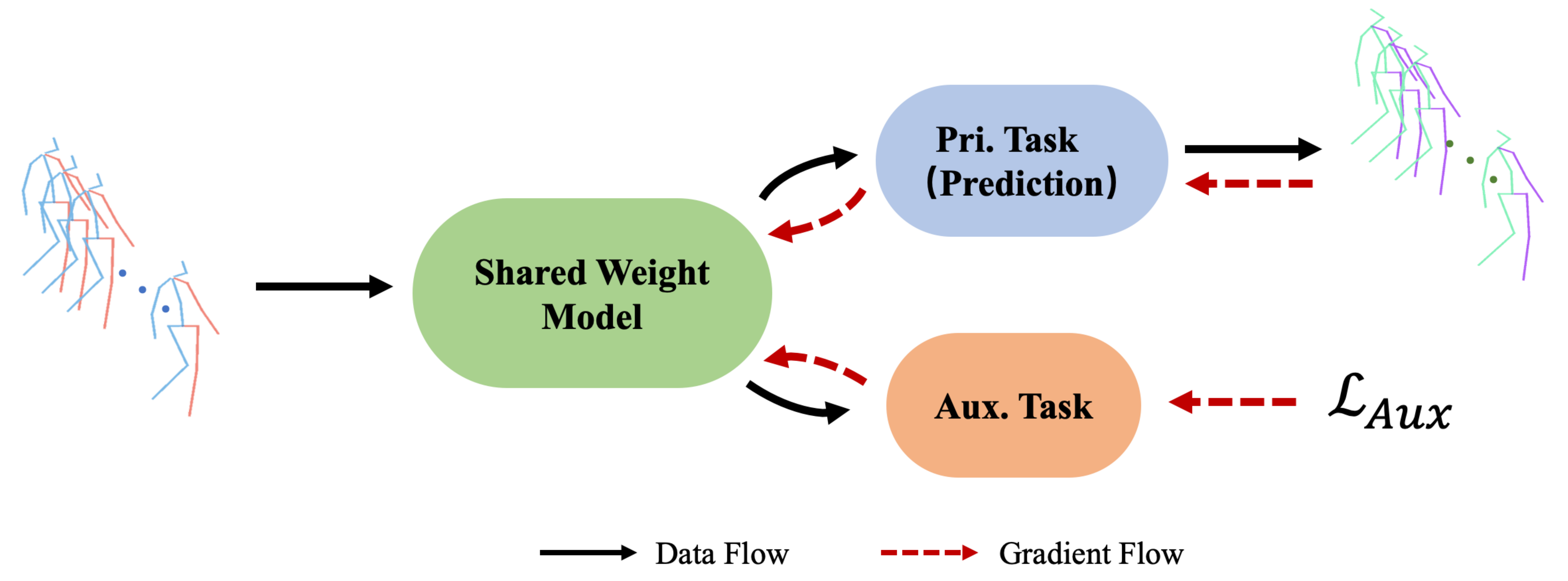}
  \caption{\textbf{Basic architecture of the auxiliary task modeling framework}, The black arrows represent the data transmission direction, while the red dashed arrows indicate the gradient backpropagation direction. After inputting the historical action sequences into the shared-weight model, the resulting features are separately fed into prediction heads for forecasting and auxiliary task heads for computing auxiliary losses, thereby enhancing the performance of the shared-weight model.}
  \label{fig:9}
\end{figure}

As shown in \hyperref[fig:9]{Fig.~\ref*{fig:9}}, in addition to the prediction task of the target, some recent work can also introduce some auxiliary tasks to strengthen the original model's ability to learn features. Auxiliary tasks play an important role in human motion prediction. They not only provide additional information but also guide feature learning and enhance the robustness of the model. By combining with prediction tasks, these auxiliary tasks effectively enhance the performance and generalization ability of motion prediction models, so they have received widespread attention in recent years. In the latest research, some work \cite{cuiAccurate3DHuman2021, xuAuxiliaryTasksBenefit2023} adopted a unified framework, integrating the tasks of predicting future postures and repairing missing values, with the repair task aiming to assist the learning of the prediction task. Xu et al.\cite{xuAuxiliaryTasksBenefit2023} also introduced a denoising task, adding random noise to the joints and then eliminating the noise by calculating the correlation between joints. Both tasks require the model to have excellent learning ability for spatio-temporal dependencies, so 
they can provide extra impetus for the model to learn spatio-temporal dependencies more comprehensively. In addition, some research \cite{liClassguidedHumanMotion2023,martinez-gonzalezPoseTransformersPOTR, liSymbioticGraphNeural2019, butepageDeepRepresentationLearning2017} also introduced category prediction tasks, that is, using the category labels of the action sequences themselves for supervised training. This method helps the network to better learn the semantic information of the sequence. Unlike other works, Li et al.\cite{liMultitaskNonAutoregressiveModel2021} not only used encoded features to predict categories but also used predicted sequences to further predict categories. The advantage of this strategy is that the predicted sequence can retain both low-level skeleton features and high-level category features. On the other hand, Li et al.\cite{liPedestrianCrossingAction2023} proposed two interesting auxiliary tasks. One involves dividing the sequence into multiple subsequences, shuffling them, and then sorting them. The other task entails shuffling the entire sequence and subsequently applying contrastive learning. This helps the model to better understand the sequential relationships and structures in the action sequences and improves the model's ability to model the sequences. In addition, Cui et al.'s\cite{cuiMetaAuxiliaryLearningAdaptive2023} research task involves shuffling sequences and determining whether they belong to the same sequence as the original sequence. Finally, we summarize these methods in \hyperref[table:6]{Table~\ref{table:6}}

\begin{table*}[ht]
\centering
\caption{Methods for introducing Auxiliary Tasks.}
\label{table:6}
\begin{tabular}{lp{10cm}llll} 
\hline
Method                                                   & Insights                                                                                                                                                                              & Venue                                   & Year & Code                                               \\ 
\hline
AuxFormer\cite{xuAuxiliaryTasksBenefit2023}              & Introducing restoration and denoising tasks to enable the model to learn a more comprehensive spatiotemporal dependency                                                                 & ICCV                                    & 2023 & No                                                 \\
MAL-Attention\cite{cuiMetaAuxiliaryLearningAdaptive2023} & Utilizing two self-supervised auxiliary tasks, namely classification and restoration, to aid in training. During testing, model parameters can be fine-tuned using both auxiliary tasks & arXiv                                   & 2023 & No                                                 \\
CGHMP\cite{liClassguidedHumanMotion2023}                 & Incorporate a prediction task to forecast motion category information                                                                                                                   & \textcolor[rgb]{0.216,0.255,0.318}{NCA} & 2023 & \href{https://github.com/cobblestones/CGHMP}{Yes}  \\
MT-GCN\cite{cuiAccurate3DHuman2021}                      & The auxiliary task is a restoration task, aimed at fixing incomplete sequences                                                                                                          & CVPR                                    & 2021 & No                                                 \\
POTR\cite{martinez-gonzalezPoseTransformersPOTR}         & When using the transformer for encoding, an additional token is introduced to learn category information and is ultimately used for classification                                      & ICCV                                    & 2021 & \href{https://github.com/idiap/potr}{Yes}          \\
mNAT\cite{liMultitaskNonAutoregressiveModel2021}         & Perform category classification on both historical sequences and prediction sequences to ensure that the predicted sequences retain essential category features                         & IEEE TIP                                & 2020 & No                                                 \\
\hline
\end{tabular}
\end{table*}

The types of auxiliary tasks are diverse and mostly belong to self-supervised tasks that can provide unsupervised learning signals. By carefully designing these auxiliary tasks, researchers can train models in the absence of labeled data, enhance model generalization capabilities and feature representation, thereby improving action prediction performance. These methods bring new possibilities to the field of human motion prediction and provide powerful insights for future research.

\begin{table*}[ht]
\centering
\caption{Methods for Modifying Prediction Approaches.}
\label{table:7}
\begin{tabular}{lp{8.8cm}llll} 
\hline
Method                                                   & Insights                                                                                                                             & Venue                                    & Year & Code                                                      \\ 
\hline
LPAttention\cite{LearningProgressiveJoint}               & Progressively predict joints from center to periphery                                                                                & \textcolor[rgb]{0.204,0.286,0.369}{ECCV} & 2020 & No                                                        \\
TrajGCN\cite{maoLearningTrajectoryDependencies2020}      & Initialize the predicted pose with the current-time pose and obtain the predicted pose through refinement                            & \textcolor[rgb]{0.204,0.286,0.369}{ICCV} & 2019 & \href{https://github.com/wei-mao-2019/LearnTrajDep}{Yes}  \\
PVRED\cite{wangPVREDPositionVelocityRecurrent2021}       & Train using quaternions                                                                                                              & IEEE TIP                                 & 2021 & No                                                        \\
PK-GCN\cite{sunOverlookedPosesActually2022}              & Simulate Priv features with Obs sequences and then perform interpolated prediction                                                   & \textcolor[rgb]{0.216,0.255,0.318}{ECCV} & 2022 & No                                                        \\
BiAttention\cite{zhaoBidirectionalTransformerGAN2023}    & Bidirectional prediction, using the predicted sequence to further predict historical sequences                                       & \textcolor[rgb]{0.216,0.255,0.318}{TOMM} & 2023 & No                                                        \\
MAL-Attention\cite{cuiMetaAuxiliaryLearningAdaptive2023} & Fine-tuned network parameters with some self-supervised tasks before prediction to achieve network parameter adaptation to test data & arXiv                                    & 2023 & No                                                        \\
Multi-GCN\cite{renMultiGraphConvolutionNetwork2023}      & Utilize offset and anchor (predicting joint position range) for prediction                                                           & arXiv                                    & 2023 & No                                                        \\
ResChunk\cite{zandMultiscaleResidualLearning2023}        & Mitigate the feature smoothing issue by calculating the offset                                                                       & arXiv                                    & 2023 & \href{https://github.com/MohsenZand/ResChunk}{Yes}        \\
KD-Former\cite{daiKDFormerKinematicDynamic2023}          & Convert exponential mapping to quaternions for training                                                                              & Pattern Recognition                      & 2023 & \href{https://github.com/wslh852/KD-Former}{Yes}          \\
LSM-GCN\cite{wangLearningSnippettoMotionProgression2023} & Predict key poses and reconstruct the remaining poses (linear interpolation) using the key poses                                     & arXiv                                    & 2023 & No                                                        \\
\hline
\end{tabular}
\end{table*}

\begin{table*}[ht]
\centering
\caption{Methods for innovating Loss Functions.}
\label{table:8}
\begin{tabular}{lp{12cm}llll} 
\hline
Method                                                        & Insights                                                                                                 & Venue                                    & Year & Code                                               \\ 
\hline
IT-GCN\cite{sunAccurateHumanMotion2023}                       & The velocity loss is applied, and the magnitude of the loss changes based on the importance of the joint & arXiv                                    & 2023 & No                                                 \\
CGHMP\cite{liClassguidedHumanMotion2023}                      & A multi-scale loss function, acceleration loss, and cross-entropy loss are used                          & NCA                                      & 2023 & \href{https://github.com/cobblestones/CGHMP}{Yes}  \\
siMLPe\cite{guoBackMLPSimple}                                 & Introducing bone length loss to narrow down the search range of predicted values                         & \textcolor[rgb]{0.216,0.255,0.318}{WACV} & 2023 & \href{https://github.com/dulucas/siMLPe}{Yes}      \\
HR-STAN\cite{medjaouriHRSTANHighResolutionSpatioTemporal2022} & The loss function uses the cosine loss of bones                                                          & CVPR                                     & 2022 & No                                                 \\
\hline
\end{tabular}
\end{table*}

\begin{figure*}
  \centering
  \includegraphics[width=1\textwidth]{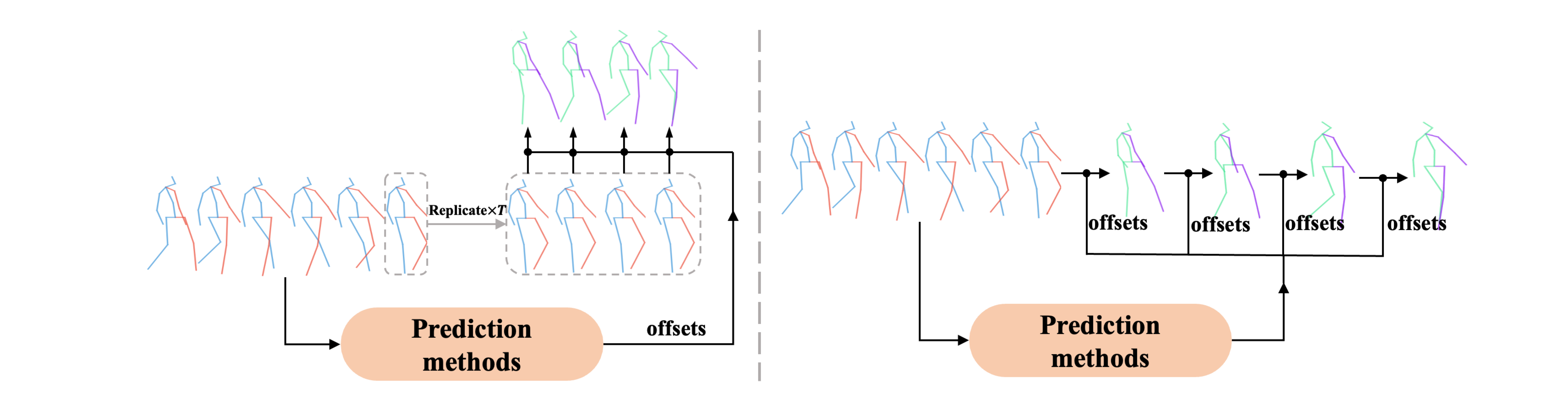}
  \caption{\textbf{Distinguishing residual prediction from offset prediction}. In the left framework, known as residual prediction, the current frame is typically replicated N times, and a single offset is generated by the model to adjust the pose positions of these N frames. On the other hand, in offset prediction, the model continuously generates multiple offsets, iteratively refining the pose based on the previous frame.}
  \label{fig:11}
\end{figure*}

\subsection{Changes in prediction approaches}
Since \cite{maoLearningTrajectoryDependencies2020}, researchers have widely adopted the method of residual prediction, because the prediction method of fine-tuning the initial sequence helps the model to better focus on the minor differences between the initial estimate and actual observation, thereby improving the adaptability to different motion patterns and changes, and achieving more accurate prediction results. However, it is worth noting that this prediction method is not immutable, and many scholars are constantly trying to innovate. For example, Wang et al.\cite{wangLearningSnippettoMotionProgression2023} believe that it is difficult to predict all frames at once, so they focus on predicting key postures and reconstruct the remaining postures through simple interpolation. The idea behind this method is that action sequences can be effectively represented by a few key postures. Sun et al.\cite{sunOverlookedPosesActually2022}, inspired by sequence interpolation algorithms, abandoned the traditional extrapolation prediction method and instead used distillation learning to simulate privileged sequence features for interpolation prediction. The uniqueness of this method lies in its flexibility and high abstraction of features.

Ren et al.\cite{renMultiGraphConvolutionNetwork2023} proposed two prediction methods. The first involves having the model predict the offset for each frame, as illustrated in \hyperref[fig:11]{Fig.~\ref*{fig:11}}, and the other is to predict the anchor points that define the joint position range. Zand et al.\cite{zandMultiscaleResidualLearning2023} similarly employ an offset-based strategy for prediction. However, they compute offsets for multiple frames simultaneously, and their paper also highlights that this approach aids in addressing the issue of over-smoothing (pattern collapse) often encountered in graph convolution methods. Cui et al.\cite{cuiMetaAuxiliaryLearningAdaptive2023} fine-tuned network parameters with some self-supervised tasks before prediction to achieve network parameter adaptation to test data. This method helps improve the generalization ability of the model and perform well in different scenarios. Dai et al.\cite{daiKDFormerKinematicDynamic2023} first predicted the angle representation of future sequences and then converted it into quaternion for training. The principle behind this strategy is that human posture may encounter deadlock and discontinuity in exponential mapping as pointed out in \cite{wangPVREDPositionVelocityRecurrent2021}, while quaternion space can effectively overcome these problems and ensure continuity and stability of posture prediction. In addition, Cai et al.\cite{LearningProgressiveJoint} proposed an interesting point of view that human motion follows the principle of motion chain, for example, the absolute displacement of the wrist is usually transmitted through the initial movement of the shoulder, followed by elbow movement. Therefore, they do not directly predict the entire human posture, but gradually predict joint trajectories from center to periphery. This progressive prediction strategy is more in line with the biomechanical characteristics of actual motion. Finally, Zhao et al.\cite{zhaoBidirectionalTransformerGAN2023} not only used forward prediction but also used predicted future sequences to infer past historical sequences in reverse, which is called backward prediction. The principle behind this method is that if predicted postures gradually tend towards average posture, backward prediction may lead to increasing errors with historical postures. Therefore, backward prediction can help alleviate the problem of predicted postures gradually tending towards average posture to some extent. We summarize the above-mentioned methods in \hyperref[table:7]{Table~\ref{table:7}}

In summary, researchers have explored a variety of innovative methods in posture prediction field to continuously improve model performance and adaptability. These methods have their own characteristics and can choose appropriate strategies according to specific problem requirements to achieve more accurate motion posture prediction.

\subsection{Loss functions}
Innovation in loss functions has always been a focus of attention in the field of deep learning. An excellent loss function can more accurately guide the core objective of the learning task during model training, thereby enhancing model performance and improving prediction accuracy. Sun et al.\cite{sunAccurateHumanMotion2023} proposed an innovative weighted L2 loss function. They cleverly introduced human dynamics information by assigning higher weights to joints with a wide range of motion and short prediction time, which to some extent improved the problem of error accumulation. In addition, the velocity loss term of this loss function also helps to prevent long-term prediction postures from becoming overly smooth. Li et al.\cite{liClassguidedHumanMotion2023} used an acceleration loss function to better describe dynamic parameters. Guo et al.\cite{guoBackMLPSimple} introduced bone length loss to narrow down the search range of posture prediction. Omar et al.\cite{medjaouriHRSTANHighResolutionSpatioTemporal2022} used the cosine similarity of the skeleton as part of the loss function. The innovation of these methods lies in their focus on different aspects of posture prediction problems, providing diversified tools to enhance model performance.

\begin{sloppypar}
In addition, Tang et al.\cite{tangLongTermHumanMotion2018} have put forth a compelling perspective, suggesting that the Mean Squared Error (MSE) loss function often focuses solely on interjoint errors, which can lead to a tendency for posture predictions to converge towards average poses without considering the continuity and consistency of sequences. Consequently, they propose a method that minimizes the Gram matrix between consecutive movements to better capture the characteristics and coherence of action sequences. This viewpoint underscores the importance of considering the characteristics of action sequences. These methods are reviewed and summarized in \hyperref[table:8]{Table~\ref{table:8}}. In summary, innovation in loss functions has played a pivotal role in the field of pose prediction, offering a rich array of options for enhancing model performance and improving the accuracy of posture prediction.
\end{sloppypar}

\subsection{Other considerations}
When discussing future research directions in the field of human motion prediction, in addition to the existing mainstream issues, there are also some niche but potentially valuable directions. First, we need to pay attention to the noise issues in the data. These noises or disturbances can damage the original motion information and lead to inaccurate predictions. It is worth noting that past research methods often did not fully discuss this issue. Wang et al.\cite{wangSpatioTemporalBranchingMotion2023} proposed an interesting point of view that each measurement data has a certain error, and between adjacent frames, the time interval is very short, so it can be reasonably assumed that the size of the error is also similar. Based on this point of view, they used the increment of adjacent frames as model input, thereby successfully reducing the impact of errors and greatly improving short-term prediction performance.

In addition, a rarely discussed issue in motion prediction is uncertainty analysis, especially uncertainty modeling methods widely used in fields such as image classification and natural language processing. Saeed et al.\cite{saadatnejadReliableHumanPose2023} introduced a method for modeling uncertainty by using uncertainty priors to inject knowledge about uncertain behavior. This method not only allows the model's capabilities to focus on more meaningful supervision directions but also reduces the number of learning parameters and improves model stability. It is worth mentioning that they also quantified the model's cognitive uncertainty by clustering and measuring the entropy of distribution, which is a promising research direction.

Another interesting direction is the initialization method of posture sequences. The traditional initialization method is achieved by replicating the current observation frame multiple times; however, this simple approach can deviate significantly from the true pose when predicting time growth. So researchers have proposed various innovative methods to improve the initialization process. For example, Sun et al.\cite{sunAccurateHumanMotion2023} alternately transformed representations between original 3D coordinate space and frequency domain space through domain conversion units, achieving iterative improvement of predicted motion. He et al.\cite{heInitialPredictionFinetuning2023} first predicted the initial human posture sequence in frequency domain space and then fine-tuned it in the original 3D coordinate space, providing a new way for improving prediction. Ma et al.\cite{maProgressivelyGeneratingBetter2022} used the prediction value of the previous stage as the initial value of the future action sequence at the current stage and performed multi-stage fine-tuning to further improve prediction accuracy.

In this section, we conducted a review of recent articles and categorized these methods according to their innovative directions, as detailed in\hyperref[table:9]{Table~\ref{table:9}}. These niche but promising research directions provide new ideas and challenges for future development in the field of human motion prediction. By deeply studying these issues, we hope to continuously improve prediction accuracy and stability, providing more reliable solutions for applications in fields such as robotics technology and medical rehabilitation.

\section{Datasets and evaluation metrics}
\subsection{Datasets}
\begin{table*}[ht]
\centering
\caption{Dataset basic parameters overview}
\label{table:9}
\begin{tabular}{>{\centering\hspace{0pt}}m{0.094\linewidth}>{\centering\hspace{0pt}}m{0.227\linewidth}>{\centering\hspace{0pt}}m{0.104\linewidth}>{\centering\hspace{0pt}}m{0.183\linewidth}>{\centering\hspace{0pt}}m{0.054\linewidth}>{\centering\hspace{0pt}}m{0.102\linewidth}>{\centering\arraybackslash\hspace{0pt}}m{0.067\linewidth}} 
\hline
Dataset & Sensors              & Number of Actions & Number of joints & FPS & Location & Year  \\ 
\hline
H36M    & 4 calibrated cameras & 15                & 32               & 25  & Indoor   & 2014  \\
CMU     & 12 infrared MX-40    & 8                 & 38               & 25  & Indoor   & 2003  \\
3DPW    & /                    & /                 & 26               & 30  & Outdoor  & 2018  \\
\hline
\end{tabular}
\end{table*}
\begin{sloppypar}
Datasets play a crucial role in the research of human motion prediction, serving as the foundation. Through reasonable dataset selection and preparation, researchers can verify the performance, generalization ability, and adaptability of models to different scenarios. In this section, we will explore a series of datasets used for human motion prediction, which provide valuable resources and challenges for research in various aspects. Here, we will not discuss datasets that have gradually fallen out of use in recent years, such as NTU RGB+D\cite{shahroudyNTURGBLarge2016} and G3D\cite{bloomG3DGamingAction2012}.
\end{sloppypar}

\textbf{Human3.6M}\cite{Human36MLarge} Human3.6M (H36M) is widely regarded as an important benchmark dataset in the field of human motion prediction. It contains about 3.6 million 3D human posture annotations and corresponding image data. This dataset covers the behaviors of 11 professional actors (including 6 males and 5 females) in 17 different scenarios, such as discussion, smoking, taking photos, and calling. To ensure fairness and consistency in experiments, researchers often choose S1, S6, S7, S8, and S9 as training data, S11 as validation set, and use S5 as test set to evaluate models on 15 action categories. It’s worth noting that when analyzing human postures, some points are often removed to avoid the influence of duplicate points and simplify calculations. For example, the hip joint that serves as a reference is often not considered, so the actual number of joints that the model needs to predict is usually 22. In addition, in data processing, for efficiency and accuracy of experiments, data is often downsampled with one frame of data extracted for analysis every 40ms.

\textbf{CMU Motion Capture}\footnote{This dataset has been made publicly accessible via http://mocap.cs.cmu.edu/} 
The CMU Mocap dataset is also a commonly used dataset captured by 12 high-precision infrared MX-40 cameras. This dataset includes 2235 video clips from 144 different participants. These participants performed various daily activities including common actions such as basketball, football, running, walking, window cleaning etc. The human posture in the original data is described by 38 joint points but in most experiments only 25 main joint points are used to keep consistency with other standard datasets (such as H36M). It’s worth noting that to standardize and unify processing these video data were downsampled to a frame rate of 25fps. Such data processing aims to ensure more accurate and consistent comparisons between different datasets.

\textbf{3DPW}\cite{vonmarcardRecoveringAccurate3D2018} 3DPW is an important real dataset for research on natural human actions in complex outdoor environments. As it covers a variety of varied and challenging scenarios this dataset has high reference value in the fields of human posture estimation and action recognition. 3DPW includes 60 video sequences totaling over 51,000 indoor and outdoor posture samples. These samples come from seven different participants who wore 18 different styles of clothing in the videos and performed a series of daily activities such as shopping, sports, hugging, talking, selfies, taking buses, playing guitar and relaxing etc. It’s worth noting that each posture in the dataset consists of 26 joints. And all video sequences were collected at a frame rate of 30fps ensuring smoothness and continuity of actions.

\textbf{AMASS}\cite{mahmoodAMASSArchiveMotion2019} The AMASS dataset is a collection of multiple motion capture datasets unified through the parameterization of SMPL+H and SMPL+X. The merged dataset comprises a total of 2992.34 minutes of motion sequences, spanning 484 distinct subjects and encompassing over 11,000 actions from all the databases. Due to its substantial size, this dataset is commonly employed as a training set to enhance model generalization. Typically, researchers use AMASS-BMLrub as the test set, while the remaining portions of the AMASS dataset are divided into training and validation sets for evaluating models on 18 joints.

Finally, We have compiled the basic parameters of several datasets in \hyperref[table:7]{Table~\ref{table:7}}. It is important to note that the sample counts may vary due to the use of different numbers of historical frames by various methods. However, it is worth mentioning that 3DPW primarily focuses on human body pose and shape, and does not include information related to action categories. Furthermore, AMASS consists of a collection of multiple subdatasets, each with different attributes such as scenes and actions. Therefore, we have refrained from specific statistical analysis in this context.

\subsection{Evaluation metrics}
The two primary evaluation metrics are Mean Angle Error(MAE) and Mean Per Joint Position Error(MPJPE). MAE is used to measure the error between the predicted pose represented in angles and the ground truth pose. However, it is important to note that angle representation is ambiguous, as different poses can have the same angle representation. This ambiguity can affect the accuracy of MAE. Consequently, in recent years, MPJPE has gradually become a more widely adopted evaluation metric for more accurately assessing the performance of pose predictions.

\textbf{Mean Angle Error(MAE)}  The MAE quantifies the error between the predicted pose, represented in angles, and the true pose. However, it is essential to be aware that angle representation can be ambiguous, potentially resulting in different poses having the same angle representation, which can impact the accuracy of MAE. Thus, in certain situations, MAE may not be suitable for evaluating pose predictions with complete accuracy.
$$
MAE=\frac1T\sum_{i=1}^T\frac1J\sum_{j=1}^J\left|\theta_{ij}^{GT}-\theta_{ij}^{Pred}\right|
$$
Here, $T$ represents the number of frames, $J$ represents the number of joints, $\theta_{ij}^{GT}$ is the angle of the $j$-th joint on the $i$-th frame of the ground truth pose, and $\theta_{ij}^{Pred}$ is the angle of the $j$-th joint on the $i$-th frame of the predicted pose.

\textbf{Mean Per Joint Position Error(MPJPE)}
MPJPE is a commonly used evaluation metric for measuring the error between predicted joint positions and actual joint positions. It calculates the average Euclidean distance between the predicted joint positions and the true joint positions, reflecting the positional accuracy of the predictions compared to the ground truth. In contrast to error measurements based on angle representation, MPJPE directly assesses the geometric shape accuracy of the pose and is therefore often considered a more accurate and reasonable evaluation metric in many cases.
$$
MPJPE=\frac1T\sum_{i=1}^T\frac1J\sum_{j=1}^J\|\mathbf{p}_{ij}^{GT}-\mathbf{p}_{ij}^{Pred}\|_2
$$

\begin{table*}[ht]
\centering
\caption{\textbf{Results on Human3.6M.} The best results are highlighted in bold, and the second best are marked by underline.}
\label{table:10}
\begin{tabular}{c|cccc|cc|l} 
\hline
\multicolumn{7}{c|}{Mean Per Joint Position Error (in millimeter)}                                             & \multicolumn{1}{c}{\multirow{2}{*}{Year}}  \\ 
\cline{1-7}
milliseconds   & 80           & 160           & 320           & 400           & 560           & 1000           & \multicolumn{1}{c}{}                       \\ 
\hline
DMAB\cite{gaoDecomposeMoreAggregate}           & 9.3          & \underline{19.7}          & \textbf{41.0} & \textbf{51.1} & \textbf{67.2} & \textbf{100.3} & 2023                                                                                                                   \\
GA-MIN\cite{zhongGeometricAlgebrabasedMultiview2023}         & 9.4          & 19.9          & 42.4          & \underline{52.2}          & \underline{68.0}          & \underline{102.2}          & 2023                                                                                                                   \\
STB-GCN\cite{wangSpatioTemporalBranchingMotion2023}        & \textbf{4.3} & \textbf{15.5} & \underline{41.8}          & 54.2          & 75.4          & 111.0          & 2023                                                                                                                   \\
EqMotion\cite{xuEqMotionEquivariantMultiagent2023}       & \underline{9.1}          & 20.1          & 43.7          & 55.0          & 73.4          & 106.9          & 2023                                                                                                                   \\
IT-GCN\cite{sunAccurateHumanMotion2023}         & \underline{9.1}          & -             & -             & 55.5          & 74.5          & 109.2          & 2023                                                                                                                   \\
GcnMlpMixer\cite{wangMixerLayerWorth2023}    & 9.4          & 21.3          & 45.8          & 56.7          & 75.2          & 108.6          & 2023                                                                                                                   \\
siMLPe\cite{guoBackMLPSimple}         & 9.6          & 21.7          & 46.3          & 57.3          & 75.7          & 109.4          & 2023                                                                                                                   \\
SANet\cite{heInitialPredictionFinetuning2023}           & 9.6          & 21.7          & 46.5          & 57.9          & 77.2          & 109.4          & 2023                                                                                                                   \\
PatternGCN\cite{tangCollaborativeMultidynamicPattern2023}     & 10.3         & 22.9          & 47.9          & 58.1          & 77.0          & 109.9          & 2023                                                                                                                   \\
CGHMP\cite{liClassguidedHumanMotion2023}           & 10.3         & 22.8          & 48.1          & 59.3          & 78.0          & 112.0          & 2023                                                                                                                   \\
LSM-GCN\cite{wangLearningSnippettoMotionProgression2023}        & 10.0        & -             & -             & 59.7          & 76.9          & 109.8          & 2023                                                                                                                   \\
DS-GCN\cite{fuLearningConstrainedDynamic2023}         & 10.4         & 23.3          & 48.8          & 59.8          & 77.8          & 111.0          & 2023                                                                                                                   \\
DANet\cite{caoDualAttentionModel2022}          & 9.8          & 21.2          & 44.0          & 54.6          & 73.3          & 107.7          & 2022                                                                                                                   \\
HR\_STAN\cite{medjaouriHRSTANHighResolutionSpatioTemporal2022}       & -            & -             & -             & 56.4          & -             & -              & 2022                                                                                                                   \\
SPGSN\cite{liSkeletonPartedGraphScattering2022}           & 10.4         & 22.3          & 47.1          & 58.3          & 77.40         & 109.6          & 2022                                                                                                                   \\
TC-CNN\cite{tangTemporalConsistencyTwostream2022}         & 9.8          & 22.6          & 48.1          & 58.4          & 76.3          & 109.6          & 2022                                                                                                                   \\
PGBIG\cite{maProgressivelyGeneratingBetter2022}          & 10.3         & 22.7          & 47.4          & 58.5          & 76.9          & 110.3          & 2022                                                                                                                   \\
MotionMixer\cite{bouaziziMotionMixerMLPbased3D2022}    & 11.0         & 23.6          & 47.8          & 59.3          & 77.8          & 111.0          & 2022                                                                                                                   \\
StaAttention\cite{uedaSpatiotemporalAggregationSkeletal2022}   & 9.2          & 21.6          & 48.6          & 61.3          & -             & -              & 2022                                                                                                                   \\
PK-GCN\cite{sunOverlookedPosesActually2022}         & 10.8         & 23.3          & 48.2          & 57.4          & 76.1          & 106.4          & 2022                                                                                                                   \\
MT-GCN\cite{cuiAccurate3DHuman2021}         & 11.0         & 22.8          & 47.9          & 58.9          & -             & 110.7          & 2022                                                                                                                   \\
MultiAttention\cite{maoMultilevelMotionAttention2021} & 10.2         & 22.2          & 46.3          & 57.3          & 75.9          & 110.1          & 2021                                                                                                                   \\
STS-GCN\cite{sofianosSpaceTimeSeparableGraphConvolutional2021}        & 11.4         & 24.8          & 51.6          & 62.9          & 81.1          & 113.8          & 2021                                                                                                                   \\
MSR-GCN\cite{dangMSRGCNMultiScaleResidual}        & 12.1         & 25.6          & 51.6          & 62.9          & 81.1          & 114.2          & 2021                                                                                                                   \\
TIM-GCN\cite{lebaillyMotionPredictionUsing2021}         & 11.4         & 24.3          & 50.4          & 60.9          & -             & -              & 2021                                                                                                                   \\
TrajectoryNet\cite{liuTrajectoryNetNewSpatiotemporal2020}  & 10.2         & 23.2          & 49.3          & 59.7          & 77.7          & 110.6          & 2020                                                                                                                   \\
Traj-GCN\cite{maoLearningTrajectoryDependencies2020}       & 12.7         & 26.1          & 52.3          & 63.5          & 81.6          & 114.3          & 2020                                                                                                                   \\
DMGNN\cite{liDynamicMultiscaleGraph2020}           & 14.7         & 28.9          & 55.5          & 67.4          & 84.0          & 129.1          & 2020                                                                                                                   \\
\hline
\end{tabular}
\end{table*}

Here, $T$ represents the number of frames, $J$ represents the number of joints, $\mathbf{p}_{ij}^{GT}$ is the 3D coordinates of the $j$-th joint on the $i$-th frame of the ground truth pose, and $\mathbf{p}_{ij}^{Pred}$ is the 3D coordinates of the $j$-th joint on the $i$-th frame of the predicted pose.

\section{Performance comparison of methods}
In the previous section, we delved into the datasets required for human motion prediction and the associated evaluation metrics. This enables us to compare current methods based on these factors. It is worth noting that previous reviews have typically employed Mean Absolute Error (MAE) as the performance metric for method comparisons. But the recent research trend is to use Mean Per Joint Position Error (MPJPE) as the primary evaluation metric because it is more accurate and reasonable compared to MAE. Therefore, in this section, we will also adopt MPJPE as the standard for assessing method performance. However, it should be pointed out that in the field of human motion prediction, there is a lack of unified evaluation standards, making it relatively challenging to comprehensively compare different methods. This underscores one of the important issues that need to be addressed in this field. In this section, we employ the most widely used evaluation approach, testing on datasets such as Human3.6M, CMU MOCAP, 3DPW, AMASS, and computing MPJPE on specific frames, as opposed to calculating the average error across all frames, as some methods do.

\begin{table*}[ht]
\centering
\caption{\textbf{Results on CMU MOCAP.} The best results are highlighted in bold, and the second best are marked by underline.}
\label{table:11}
\begin{tabular}{c|cccc|cc|l} 
\hline
\multicolumn{7}{c|}{Mean Per Joint Position Error (in millimeter)}                                          & \multicolumn{1}{c}{\multirow{2}{*}{Year}}  \\ 
\cline{1-7}
milliseconds & 80           & 160           & 320           & 400           & 560           & 1000          & \multicolumn{1}{c}{}                       \\ 
\hline
DMAB\cite{gaoDecomposeMoreAggregate}         & 6.4  & 13.9 & 27.9 & 36.0 & 50.1 & 75.4 & 2023                                       \\
GA-MIN\cite{zhongGeometricAlgebrabasedMultiview2023}   & 8.7  & 15.4 & 31.9 & -    & \underline{40.1} & 84.5 & 2023                                       \\
SANet\cite{heInitialPredictionFinetuning2023}   & 7.1  & 13.2 & 26.8 & 34.1 & 47.6 & \underline{72.4} & 2023                                       \\
PatternGCN\cite{tangCollaborativeMultidynamicPattern2023} & 8.4  & 14.5 & 30.6 & 39.7 & -    & 91.3 & 2023                                       \\
CGHMP\cite{liClassguidedHumanMotion2023}   & 9.0  & 16.1 & 31.6 & 40.1 & -    & -    & 2023                                       \\
LSM-GCN\cite{wangLearningSnippettoMotionProgression2023} & 10.6 & 17.9 & 33.9 & 41.2 & 56.6 & 88.0 & 2023                                       \\
DS-GCN\cite{fuLearningConstrainedDynamic2023}   & 7.3  & 13.9 & 28.4 & 35.9 & 50.1 & 80.0 & 2023                                       \\
SPOTR\cite{nargundSPOTRSpatiotemporalPose2023} & 16.2 & 21.9 & 34.6 & 42.4 & -    & -    & 2023                                       \\
SC-GCN\cite{chenSpatiotemporalConsistencyLearning2023} & 8.6  & 16.9 & 31.6 & 41.2 & -    & 82.5 & 2023                                       \\
FMS-AM\cite{fernandoRememberingWhatImportant2023}   & 6.1  & \textbf{10.3} & \textbf{22.6} & \underline{29.6} & \textbf{32.4} & \textbf{52.5} & 2023                                       \\
SPGSN\cite{liSkeletonPartedGraphScattering2022}   & 8.3  & 14.8 & 28.6 & 37.0 & -    & 77.8 & 2022                                       \\
TC-CNN\cite{tangTemporalConsistencyTwostream2022}   & 8.2  & 15.1 & 32.8 & 43.0 & -    & 92.6 & 2022                                       \\
PGBIG\cite{maProgressivelyGeneratingBetter2022}   & 7.6  & 14.3 & 29   & 36.6 & 50.9 & 80.1 & 2022                                       \\
StaAttention\cite{uedaSpatiotemporalAggregationSkeletal2022} & \textbf{5.1} & 11.6 & \underline{23.4} & \textbf{29.4} & -    & -    & 2022                                       \\
PK-GCN\cite{sunOverlookedPosesActually2022}   & 9.4  & 17.1 & 32.8 & 40.3 & 52.2 & 79.3 & 2022                                       \\
TC-GCN\cite{tangTemporalConsistencyTwostream2022}   & 6.6  & 12.4 & 26.8 & 36.3 & 51.8 & 79.8 & 2022                                       \\
TIM-GCN\cite{lebaillyMotionPredictionUsing2021}   & 10.8 & 19.5 & 36.9 & 46.2 & -    & 95.7 & 2021                                       \\
MSR-GCN\cite{dangMSRGCNMultiScaleResidual}   & \underline{5.5}  & \underline{11.1} & 25.1 & 32.5 & -    & -    & 2021                                       \\
LPAttention\cite{LearningProgressiveJoint} & 9.8  & 17.6 & 35.7 & 45.1 & -    & 93.2 & 2020                                       \\
Traj-GCN\cite{maoLearningTrajectoryDependencies2020}   & 11.5 & 20.4 & 37.8 & 46.8 & -    & 96.5 & 2020                                       \\
DMGNN\cite{liDynamicMultiscaleGraph2020}   & 12.7 & 23.3 & 42.0 & 51.1 & 68.9 & 107.0 & 2020                                       \\
\hline
\end{tabular}
\end{table*}

\textbf{Human3.6M} is the most commonly used dataset, and every method is typically evaluated on this dataset. Authors usually assess performance at several time frames: 80ms, 160ms, 320ms, 400ms, 560ms, and 1000ms. Predictions with a duration of less than 500ms are referred to as short-term predictions, while those exceeding 500ms are considered long-term predictions. Some experimental results for various methods are shown in Table \ref{table:10} . It can be seen that the STB-GCN\cite{wangSpatioTemporalBranchingMotion2023} utilizes incremental inputs to reduce the effect of errors at two time points, 80 milliseconds and 160 milliseconds, which is indeed an effective strategy. However, it is worth noting that the incremental input lacks explicit semantic information, which may lead to poor performance of STB-GCN in long-term prediction. Over time, i.e., after 160 milliseconds, the DMAB\cite{gaoDecomposeMoreAggregate} model emerges as the best performing model. This is due to DMAB's ability to extract features from different spectra, and this multi-view frequency representation enhances the capture of the spectral diversity of body movements. 

As a result, DMAB performs best among the various models in terms of long-term prediction. The remaining models, such as GA-MIN, leverage the concept of multivectors from geometric algebra to restructure features from the perspective of spectral channels, creating a more robust graph feature representation. EqMotion considers equivariance, enabling the model to learn feature representations related to geometric transformations. IT-GCN iterates between the frequency domain and spatial domain to refine the prediction of human motion. These models have all demonstrated promising performance on the Human3.6M dataset. Hence, while incremental input significantly improves short-term prediction accuracy, enhancing features from a spectral standpoint appears to be a viable approach for enhancing long-term prediction performance. It's important to note that all data is sourced from the original research papers.

\textbf{CMU MOCAP} is consistent with Human3.6M, evaluating models' performance at time frames of 80ms, 160ms, 320ms, 400ms, 560ms, and 1000ms. Experimental results are presented in Table \ref{table:11}. StaAttention\cite{uedaSpatiotemporalAggregationSkeletal2022} exhibits excellent performance at 80 ms and 400 ms time points. It establishes a network based on attention mechanisms that effectively aggregate motion sequence data in both time and space, adapting to variations in different human motion patterns while accurately representing kinematic constraints. However, at other time points, FMS-AM\cite{fernandoRememberingWhatImportant2023} excels. FMS-AM has the capability to store knowledge specific to individual subjects and particular actions. When encountering similar sample data, it can rapidly retrieve relevant information from memory, significantly enhancing prediction accuracy. The methods MSR-GCN, GA-MIN, SA-Net, and others have achieved remarkable performance improvements in various aspects, including the utilization of multi-scale features, enhanced graph feature representations, and improved initialization methods. These factors have significantly enhanced the model's performance.

\textbf{3DPW} dataset consists of real-world data from complex outdoor environments, presenting a higher level of difficulty. Researchers typically evaluate models at time frames of 200ms, 400ms, 600ms, 800ms, and 1000ms. Experimental results are summarized in Table \ref{table:12}. In this dataset, DS-GCN\cite{fuLearningConstrainedDynamic2023} outperformed until 600 milliseconds, but after 600 milliseconds, GA-MIN\cite{zhongGeometricAlgebrabasedMultiview2023} successfully surpassed it. DS-GCN employs a generic and dynamically adjustable adjacency matrix representation, allowing for dynamic adjustments based on different input samples. In contrast, GA-MIN enhances graph convolution through the utilization of the multivector concept from geometric algebra. Some work also train models on the AMASS dataset and test them on AMASS-BMLrub and 3DPW datasets. In this testing scenario, where the training and testing data come from different datasets, the differences between the training and testing data are much greater, making generalization more challenging. Experimental results for this setup are shown in Table \ref{table:13}.

\begin{table*}[ht]
\centering
\caption{\textbf{Results on 3DPW}. The best results are highlighted in bold, and the second best are marked by underline.}
\label{table:12}
\begin{tabular}{c|ccccc|c} 
\hline
\multicolumn{6}{c|}{Mean Per Joint Position Error (in millimeter)}                           & \multicolumn{1}{l}{\multirow{2}{*}{Year}}  \\ 
\cline{1-6}
milliseconds & 200           & 400           & 600           & 800           & 1000          & \multicolumn{1}{l}{}                       \\ 
\hline
DMAB\cite{gaoDecomposeMoreAggregate}         & \underline{26.1}          & \underline{54.2}          & 72.3          & 87.2          & \underline{94.5}          & 2023                                       \\
GA-MIN\cite{zhongGeometricAlgebrabasedMultiview2023}       & 27.2          & 57.1          & \underline{71.2}          & \textbf{82.1} & \textbf{90.6} & 2023                                       \\
STB-GCN\cite{wangSpatioTemporalBranchingMotion2023}      & 27.6          & 65.1          & 88.6          & 101.4         & 109.8         & 2023                                       \\
PatternGCN\cite{tangCollaborativeMultidynamicPattern2023}   & 29.5          & 58            & 84.7          & 103.1         & 109.3         & 2023                                       \\
LSM-GCN\cite{wangLearningSnippettoMotionProgression2023}      & 35.6          & 66.3          & 85.7          & 99.2          & 106.2         & 2023                                       \\
DS-GCN\cite{fuLearningConstrainedDynamic2023}       & \textbf{24.5} & \textbf{49.9} & \textbf{69.9} & \underline{85.2}          & 96.3          & 2023                                       \\
DANet\cite{caoDualAttentionModel2022}        & 37.0          & 61.9          & 76.7          & 86.2          & 95.2          & 2022                                       \\
SPGSN\cite{liSkeletonPartedGraphScattering2022}        & 32.9          & 64.5          & 91.6          & 104.0         & 111.1         & 2022                                       \\
TC-CNN\cite{tangTemporalConsistencyTwostream2022}       & 28.4          & 59.1          & 84.6          & 100           & 108.2         & 2022                                       \\
PGBIG\cite{maProgressivelyGeneratingBetter2022}        & 29.3          & 58.3          & 79.8          & 94.4          & 104.1         & 2022                                       \\
PK-GCN\cite{sunOverlookedPosesActually2022}       & 34.8          & 66.2          & 88.1          & 104.3         & 114.2         & 2022                                       \\
Traj-GCN\cite{maoLearningTrajectoryDependencies2020}     & 35.6          & 67.8          & 90.6          & 106.9         & 117.8         & 2020                                       \\
DMGNN\cite{liDynamicMultiscaleGraph2020}        & 37.3          & 67.8          & 94.5          & 109.7         & 123.6         & 2020                                       \\
LDR-GCN\cite{cuiLearningDynamicRelationships2020}      & 33.9          & 57.4          & 84.6          & 95.2          & 109.1         & 2020                                       \\
\hline
\end{tabular}
\end{table*}

\begin{table*}[ht]
\centering
\footnotesize
\caption{\textbf{Results on AMASS and 3DPW.} The best results are highlighted in bold, and the second best are marked by underline.}
\label{table:13}
\begin{tabular}{c|cccccccc||cccccccc} 
\hline
\vcell{}         & \multicolumn{8}{c||}{\vcell{AMASS}}                    & \multicolumn{8}{c}{\vcell{3DPW}}                       \\[-\rowheight]
\printcellbottom & \multicolumn{8}{c||}{\printcellmiddle}                 & \multicolumn{8}{c}{\printcellmiddle}                   \\
milliseconds     & 80   & 160  & 320  & 400  & 560  & 720  & 880  & 1000  & 80   & 160  & 320  & 400  & 560  & 720  & 880  & 1000  \\ 
\hline
IT-GCN\cite{sunAccurateHumanMotion2023}           & \underline{9.8}  & 17.8 & 31.8 & 37.9 & 47.8 & 55.3 & 60.7 & 65..4 & 11.6 & 20.8 & 36.4 & 42.8 & 53.3 & 61   & 66.9 & 71    \\
GcnMlpMixer\cite{wangMixerLayerWorth2023}      & 10.4 & 19.1 & 38.9 & 49.2 & 50.1 & 56.6 & 61.9 & 65.4  & 12   & 22.1 & 37.8 & 44.1 & 54.5 & 62.1 & 68   & 72.3  \\
siMLPe\cite{guoBackMLPSimple}           & 10.8 & 19.6 & 34.3 & 40.5 & 50.5 & 57.3 & 62.4 & 65.7  & 12.1 & 22.1 & 38.1 & 44.5 & 54.9 & 62.4 & 68.2 & 72.2  \\
HR-STAN\cite{medjaouriHRSTANHighResolutionSpatioTemporal2022}          & -    & -    & -    & 35.3 & -    & -    & -    & -     & -    & -    & -    & -    & -    & -    & -    & -     \\
MotionMixer\cite{bouaziziMotionMixerMLPbased3D2022}      & \textbf{6.6}  & \textbf{10.3} & \textbf{18}   & \textbf{21.9} & \textbf{28.8} & \textbf{33.6} & \textbf{38.8} & \textbf{41.6}  & \textbf{7.4}  & \textbf{11.4} & \underline{19.3} & \textbf{22.8} & \underline{29.3} & \textbf{34.6} & \textbf{39}   & \underline{42.1}  \\
STG-GCN\cite{zhongSpatioTemporalGatingAdjacencyGCN2022}          & 10.0 & \underline{11.9} & \underline{20.1} & \underline{24.0} & \underline{30.4} & -    & -    & \underline{43.1}  & \underline{8.4}  & \underline{11.9} & \textbf{18.7} & \underline{23.6} & \textbf{29.1} & -    & -    & \textbf{39.9}  \\
STS-GCN\cite{sofianosSpaceTimeSeparableGraphConvolutional2021}          & 10.0 & 12.5 & 21.8 & 24.5 & 31.9 & \underline{38.1} & \underline{42.7} & 45.5  & 8.6  & 12.8 & 21   & 24.5 & 30.4  & \underline{35.7} & \underline{39.6} & 42.3  \\
\hline
\end{tabular}
\end{table*}

Based on the performance across multiple datasets, the following conclusions can be drawn:
\begin{enumerate}[label={\arabic*)}, itemsep=0pt, topsep=5pt, partopsep=0pt, parsep=0pt]
	\item A more complex model does not necessarily yield better results in human motion prediction. Simple models, such as siMLPe\cite{guoBackMLPSimple}, can achieve excellent performance. Hence, it is possible to model human motion prediction tasks in a completely different and simpler manner without the explicit fusion of spatial and temporal information.
	\item Multiscale approaches are crucial for robust feature extraction in human motion prediction.
	\item Properly extracting frequency-domain spatial features can enhance a model's performance in human motion prediction.
	\item Due to the complexity of the 3DPW\cite{vonmarcardRecoveringAccurate3D2018} dataset, sample-specific models can achieve superior performance. This also underscores the need for models to possess sample-specific and task-specific capabilities for better real-world applications.
\end{enumerate}

\section{Challenges}
In recent years, significant progress has been made in the field of human motion prediction. Against this backdrop, Part 4 of this article discusses in detail the main innovative directions mentioned in recent papers and predicts the future development trends of these directions. This section will delve deeper into summarizing and looking forward to the possible development challenges of human motion prediction.

\subsection{Enhanced spatiotemporal dependency extraction}
Although existing research has made some progress in this direction, there are still certain challenges in extracting spatio-temporal dependencies. For example, when predicting some more complex actions, such as ‘greeting’, ‘purchases’, ‘sittingdown’, ‘walkingdog’, the prediction results of existing methods often have obvious deviations. In addition, in long-term predictions, the problem of postures tending towards average posture cannot be ignored, which also implies its limitations in extracting spatio-temporal dependencies from historical sequences. Therefore, deepening the study and improving the model’s ability in this aspect will be an important direction for future research.

\subsection{Innovation in auxiliary tasks}
As the methods of adjusting model structures gradually become relatively limited, auxiliary tasks have emerged as a new research method. These tasks aim to provide additional learning signals to help the main model capture and learn spatio-temporal dependencies more comprehensively. However, although auxiliary tasks provide a new learning direction for the model, their research is still in its infancy, and the forms of related methods and tasks need to be further explored. How to effectively design and integrate auxiliary tasks to better enhance model performance will be an important topic for future research.

\subsection{Exploring more reasonable loss functions}
Traditionally, loss functions for human motion prediction mainly rely on L2 loss, velocity loss, bone length loss, etc. However, these objective functions mostly only measure the distance between predicted action sequences and real action sequences, ignoring the semantic information of human actions therein. This simplified measurement method may ignore the true rationality of predicted actions, which we believe is one of the key factors affecting long-term prediction quality. Therefore, future research should pay more attention to developing loss functions that are more consistent with human action semantics.

\subsection{More comprehensive datasets}
The datasets currently used are relatively limited in scenes and types of actions and may only cover a small part of real-world situations. In order to better capture and understand the diversity and complexity in the real world, future research should strive to build more comprehensive and rich datasets. This means that datasets not only need to cover more scenes and types of actions but also consider changes in human form under different situations.

With continuous technological advancements, there is tremendous research potential in the field of human motion prediction. We look forward to future researchers leveraging these development trends to propose more innovative methods and applications, injecting more value into fields such as autonomous driving, virtual reality, medical health etc.

\section{Conclusion}
This article provides a comprehensive overview of the field of human motion prediction, focusing on several key innovative directions. We delve into recent significant advancements as well as the challenges that remain unaddressed. By systematically reviewing existing methods, we aim to provide readers with a profound understanding of this field. Furthermore, we introduce common datasets and evaluation metrics in this field, establishing a robust foundation for research endeavors. In this review, we emphasize the abundant potential within the realm of human motion prediction. With continuous technological advancements, we anticipate researchers harnessing these emerging trends to propose more innovative approaches and applications, thereby adding further value to this field. Additionally, we put forth four potential development challenges, offering valuable insights for both academia and industry practitioners. These directions are poised to address current challenges and propel the further development of human motion prediction.


\section*{Acknowledgement}
\begin{sloppypar}
This work is funded by National Natural Science Foundation of China, grant number: 62002215; This work is partly funded by Shanghai Pujiang Program (No. 20PJ1404400)
\end{sloppypar}

\bibliographystyle{elsarticle-num} 
\bibliography{cite.bib}





\end{document}